\pdfoutput=1

\documentclass[11pt]{article}

\usepackage[preprint]{acl}

\usepackage{times}
\usepackage{latexsym}

\usepackage{placeins}
\usepackage{multirow}
\usepackage{arydshln}
\usepackage{graphicx}
\usepackage{paralist}
\usepackage{amssymb}
\usepackage{amsmath}
\usepackage{pifont}
\usepackage{xcolor}
\usepackage{bbm}
\usepackage{listings}
\definecolor{codegreen}{rgb}{0,0.6,0}
\definecolor{codegray}{rgb}{0.5,0.5,0.5}
\definecolor{codepurple}{rgb}{0.58,0,0.82}
\definecolor{backcolour}{rgb}{0.95,0.95,0.92}
\definecolor{lavender}{rgb}{0.9, 0.9, 0.98}
\definecolor{mintgreen}{rgb}{0.6, 1, 0.6}
\definecolor{peach}{rgb}{1, 0.89, 0.71}
\definecolor{skyblue}{rgb}{0.53, 0.81, 0.92}
\lstdefinestyle{mystyle}{
    backgroundcolor=\color{backcolour},   
    commentstyle=\color{codegreen},
    keywordstyle=\color{magenta},
    numberstyle=\tiny\color{codegray},
    stringstyle=\color{codepurple},
    basicstyle=\ttfamily\footnotesize,
    breakatwhitespace=false,         
    breaklines=true,                 
    captionpos=b,                    
    keepspaces=true,                 
    showspaces=false,                
    showstringspaces=false,
    showtabs=false,                  
    tabsize=2
}

\usepackage{pifont}

\usepackage{cleveref}
\crefformat{section}{\S#2#1#3} 
\crefformat{subsection}{\S#2#1#3}
\crefformat{subsubsection}{\S#2#1#3}

\usepackage[T1]{fontenc}

\usepackage[utf8]{inputenc}

\usepackage{microtype}

\usepackage{inconsolata}

\title{\textsc{KPEval}: Towards Fine-Grained Semantic-Based Keyphrase Evaluation}

\author{
Di Wu, Da Yin, Kai-Wei Chang \\
University of California, Los Angeles \\ 
\texttt{\{diwu,da.yin,kwchang\}@cs.ucla.edu} 
}

\newcommand{\cmark}{\ding{52}}

\begin{document}
\maketitle

\begin{abstract}
    Despite the significant advancements in keyphrase extraction and keyphrase generation methods, the predominant approach for evaluation mainly relies on exact matching with human references. This scheme fails to recognize systems that generate keyphrases semantically equivalent to the references or diverse keyphrases that carry practical utility. To better assess the capability of keyphrase systems, we propose \textsc{KPEval}, a comprehensive evaluation framework consisting of four critical aspects: reference agreement, faithfulness, diversity, and utility. For each aspect, we design semantic-based metrics to reflect the evaluation objectives. Meta-evaluation studies demonstrate that our evaluation strategy correlates better with human preferences compared to a range of previously proposed metrics. Using \textsc{KPEval}, we re-evaluate 23 keyphrase systems and discover that (1) established model comparison results have blind-spots especially when considering reference-free evaluation; (2) large language models are underestimated by prior evaluation works; and (3) there is no single best model that can excel in all the aspects. 
\end{abstract}

\section{Introduction}

\begin{figure*}[t!]
\vspace{-5mm}
\includegraphics[width=\textwidth]{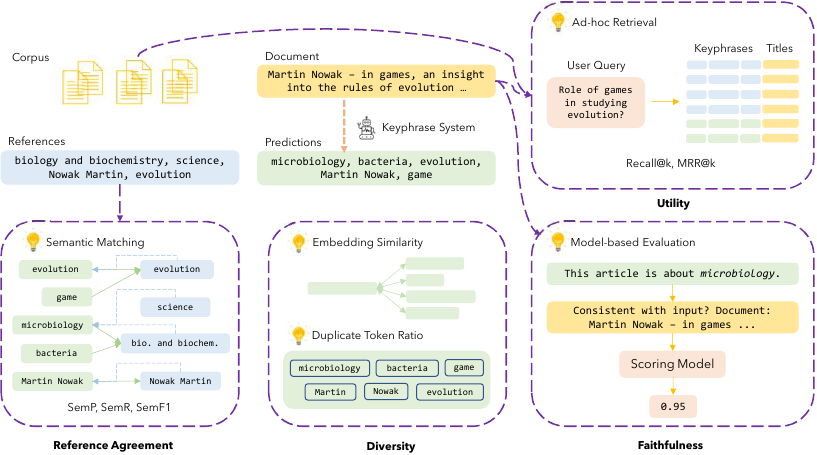}
\vspace{-6mm}
\caption{An illustration of the proposed keyphrase evaluation framework. \textsc{KPEval} evaluates keyphrase systems on four crucial properties and incorporates semantic-based metrics for more accurate assessment.}
\label{main-framework}
\vspace{-3mm}
\end{figure*}

Building automated keyphrase prediction systems has been a long-lasting research interest of Information Retrieval (IR) and Natural Language Processing (NLP) \citep{witten1999kea, hulth-2003-improved, meng-etal-2017-deep}. While a large number of keyphrase prediction systems have been proposed, the majority of them are assessed using a simplistic method: comparing the stemmed predictions with human references for exact matches. An extensive review of 76 recent keyphrase extraction and generation papers published in major conferences reveals a predominant reliance on exact matching, with 75 of 76 papers employing it and 51 treating it as the sole evaluation criterion (Appendix \ref{kp-paper-eval-survey}).

This over-reliance brings two major concerns. First, it has been established that the \textit{evaluation accuracy} of exact matching is inadequate \citep{zesch-gurevych-2009-approximate}. Although a number of heuristics are proposed to relax the matching criteria \citep{zesch-gurevych-2009-approximate, kim-etal-2010-evaluating, luo-etal-2021-keyphrase-generation, koto-etal-2022-lipkey} or enrich the label set \citep{chan-etal-2019-neural}, they still struggle to accurately capture \textit{phrase semantics} and have not been validated by systematic meta-evaluation. Second, solely relying on reference-based evaluation is an \textit{incomplete} strategy that overlooks critical aspects such as diversity of the predicted keyphrases \citep{bahuleyan-el-asri-2020-diverse} or their utility in practical applications of keyphrase systems such as indexing for IR applications \citep{boudin-gallina-2021-redefining}.

In this paper, we undertake a systematic approach to advance keyphrase evaluation. For reference-based evaluation, we propose a phrase-level semantic matching metric with a high quality embedding trained on large-scale keyphrase data. Based on the human evaluation corpus annotated on KP20k \citep{meng-etal-2017-deep} and KPTimes \citep{gallina-etal-2019-kptimes}, the meta-evaluation on five keyphrase systems shows that our metric significantly outperforms existing metrics by more than 0.15 absolute points in Kendall's Tau. By contrast, many proposed improvements to exact matching surprisingly fail to improve its human agreement  (\cref{meta-evaluation}). Further analyses reveal that the proposed metric exhibits enhanced stability under the label variations commonly present in the keyphrase annotations.

Next, we move beyond reference agreement and holistically consider the desiderata for evaluating keyphrase systems. Three crucial aspects are introduced: (1) \textit{faithfulness}, whether the predictions are grounded to the document (\cref{nat-faith-methods}); (2) \textit{diversity}, whether the predictions represent distinct concepts (\cref{diversity-def}); and (3) \textit{utility} for downstream IR applications (\cref{utility-def}). To accurately assess each aspect, we propose semantically-oriented metric designs including embedding similarity, model-based consistency evaluation, and dense retrieval. 

Together, these aspects and metrics form \textsc{KPEval}, a \textit{fine-grained semantic-based} keyphrase evaluation framework (Figure \ref{main-framework}). In \cref{re-eval}, we employ \textsc{KPEval} to evaluate 23 keyphrase systems, producing intriguing insights: 

\begin{compactenum}
    \item \textsc{KPEval} uncovers \textit{blind-spots} in established model comparisons, such as the actual superiority of ExHiRD-h \citep{chen-etal-2020-exclusive} in many aspects as well as a common difficulty to outperform a baseline in all the aspects.
    \item We find that \textit{large language models} (LLMs), particularly GPT-3.5 \citep{ouyang2022training}, exhibit remarkable performance compared to current state-of-the-art keyphrase generation and extraction models. Our results challenge existing conclusions and lead to a reconsideration of using LLMs as keyphrase systems. 
    \item Finally, \textsc{KPEval}'s four aspects test distinct abilities at which different models excel, suggesting the importance of aligning evaluation with the diverse needs of real applications.  
\end{compactenum}

In summary, \textsc{KPEval} establishes a new standard for keyphrase evaluation by advancing reference-based evaluation accuracy and aligning model development with application values via holistic reference-free evaluation. To facilitate future studies, the implementation is released as a toolkit along with the meta-evaluation annotations at \url{https://github.com/uclanlp/KPEval}.

\section{Related Work}
In this section, we review the relevant literature on evaluating keyphrase systems.

\paragraph{Reference-based evaluation} The major metrics for evaluating keyphrase systems are precision, recall, and F1 based on exact-match between the stemmed predictions and references \citep{mihalcea-tarau-2004-textrank, meng-etal-2017-deep, yuan-etal-2020-one}. This method indiscriminately penalizes unmatched predictions, including synonyms or parent/child concepts of the reference. Later works attempt to improve the metric by \textit{relaxing the matching criterion}. \citet{zesch-gurevych-2009-approximate} propose to use R-precision with approximate matching, tolerating a prediction to be a substring of a reference and vice versa. \citet{kim-etal-2010-evaluating} employ n-gram matching metrics such as BLEU \citep{papineni-etal-2002-bleu} and Rouge \citep{lin-2004-rouge}. \citet{chan-etal-2019-neural} expand the references with name variations. \citet{luo-etal-2021-keyphrase-generation} propose a fine-grained score that combines token-level matching, edit distance, and duplication penalty. \citet{koto-etal-2022-lipkey} and \citet{glazkova2022applying} use the semantic-based BertScore \citep{bert-score} with predictions and references concatenated into two strings. 

Meanwhile, ranking-based metrics such as Mean Reciprocal Rank, mean Averaged Precision, and Normalized Discounted Cumulative Gain are introduced to evaluate the ranking provided by keyphrase extraction models \citep{florescu-caragea-2017-positionrank, boudin-2018-unsupervised, kim-etal-2021-structure}. These metrics also compute exact matching to the references during their evaluation. 

\paragraph{Reference-free evaluation} Directly evaluating keyphrase predictions without references is less common. Early studies conduct human evaluation \citep{barker2000using, matsuo2004keyword}. Later work evaluates the predictions' utility in applications such as retrieval \citep{Bracewell2005MultilingualSD, boudin-gallina-2021-redefining} or summarization \citep{litvak-last-2008-graph}. \citet{bahuleyan-el-asri-2020-diverse} conduct reference-free evaluation of the predictions' diversity. 

\paragraph{Meta-evaluation} 
Meta-evaluation studies that compare keyphrase metrics with human evaluations have been limited in scope, with a focus on reference-based evaluation. \citet{kim-etal-2010-evaluating} compare five lexical matching metrics and concluded that R-precision has the highest Spearman correlation with human judgments. \citet{bougouin2016termith} annotated a meta-evaluation corpus with 400 documents in French, evaluating 3 keyphrase models on  "appropriateness" and "silence", approximately corresponding to precision and false negative rate. 

\paragraph{Discussion} Building upon existing literature, this work systematically rethinks the goals of keyphrase evaluation and advances the evaluation methodology. We introduce \textsc{KPEval}, a holistic evaluation framework encompassing four key aspects (\cref{section-framework}). \textsc{KPEval} incorporates semantic-based metrics validated via rigorous meta-evaluation (\cref{meta-evaluation} and \cref{nat-faith-methods}). Finally, we conduct a large-scale evaluation of 21 keyphrase systems and offer novel insights into existing model comparisons and LLMs (\cref{re-eval}).

\section{Background}
This section formulates the keyphrase prediction and evaluation tasks and outlines the scope of study. 

\subsection{Keyphrase Prediction}

We denote an instance of keyphrase prediction as a tuple \textbf{$(\mathcal{X},\mathcal{Y})$}, where $\mathcal{X}$ represents an input document and $\mathcal{Y}=\{y_1,...,y_n\}$ is a set of $n$ reference keyphrases provided by humans. Each $y_i$ is categorized as a \textit{present keyphrase} if it corresponds to contiguous word sequences in $\mathcal{X}$ after stemming, or an \textit{absent keyphrase} if it does not. Keyphrase generation (KPG) assumes $\mathcal{Y}$ to include both present and absent keyphrases, whereas keyphrase extraction (KPE) only allows present keyphrases in $\mathcal{Y}$.

\subsection{Keyphrase Evaluation}
\label{problem-formulation}
The \textit{keyphrase evaluation} process can be viewed as mapping a 4-element tuple $(\mathcal{X}, \mathcal{Y}, \mathcal{P}, \mathcal{C})$ to a real number via a function $f$. $\mathcal{P}=\{p_1,...,p_m\}$ is a \textit{set} of $m$ predictions made by a model $\mathcal{M}$ on $\mathcal{X}$. Different from the commonly followed works \cite{meng-etal-2017-deep,yuan-etal-2020-one}, we do not distinguish between present and absent keyphrases. This enables matching a predicted keyphrase to any semantically relevant reference, and vice versa.

$\mathcal{C}$ is a corpus that represents the domain of interest, which is an important factor in assessing keyphrase quality. For example, "sports games" may be informative in a general news domain but less so in the specialized domain of basketball news\footnote{Check \citet{tomokiyo-hurst-2003-language} for more examples.}. In this paper, $\mathcal{C}$ will play a crucial role in evaluating the utility of keyphrases for facilitating ad-hoc in-domain document retrieval (\cref{utility-def}).

\subsection{Evaluation Scope}

\paragraph{Models} This paper covers 21 representative, strong, and diverse keyphrase prediction models spanning three categories: (1) KPE models, (2) KPG models, and (3) large language models (LLMs) and APIs. We aim to include highly cited (up to February 2024) models such as MultipartiteRank
(\citet{boudin-2018-unsupervised}, 219 citations), CatSeq (\citet{yuan-etal-2020-one}, 92 citations), and SetTrans (\citet{ye-etal-2021-one2set}, 65 citations). We provide introductions and implementation details in Appendix \ref{kp-system-impl}. 

\paragraph{Datasets} We test on two datasets throughout the paper: (1) KP20k \citep{meng-etal-2017-deep} that features 20k Computer Science papers with keyphrases extracted from the paper metadata and (2) KPTimes \citep{gallina-etal-2019-kptimes} that provides 10k news documents paired with keyphrases assigned by expert editors. The two datasets are selected due to their large training sets (500k for KP20k and 250k for KPTimes) and their wide usage in keyphrase research. As such, the models' performance is easier for the community to relate to and the reproduction correctness can be verified more easily. 

\section{\textsc{KPEval}: Evaluation Aspects}
\label{section-framework}

We introduce \textsc{KPEval}, a fine-grained framework for keyphrase evaluation. \textsc{KPEval} posits to evaluate $\mathcal{P}$'s quality across four crucial aspects:

\begin{compactenum}
    \item \textbf{{Reference Agreement}}: Evaluates the extent to which $\mathcal{P}$ aligns with human-annotated $\mathcal{Y}$.
    \item \textbf{{Faithfulness}}: Determines whether each $p_i$ in $\mathcal{P}$ is semantically grounded in $\mathcal{X}$.
    \item \textbf{{Diversity}}: Assesses whether $\mathcal{P}$ includes diverse keyphrases with minimal repetitions.
    \item \textbf{{Utility}}:  Measures the potential of $\mathcal{P}$ to enhance downstream applications, such as document indexing for improved IR performance.
\end{compactenum}

Table \ref{tab:kp-sys-property-assumptions} outlines the assumptions of the evaluated aspects: whether they are calculated on a set of phrases and whether $\mathcal{X}$, $\mathcal{Y}$, or $\mathcal{C}$ is needed for evaluation. By design, these aspects have deep groundings in the previous literature. Faithfulness and reference agreement can be seen as different definitions of informativeness: the former enforces the information of $p_i$ to be contained in $\mathcal{X}$, while the latter measures $\mathcal{P}$'s coverage of $\mathcal{X}$'s salient information with respect to a background domain \citep{tomokiyo-hurst-2003-language}. Diversity \citep{bahuleyan-el-asri-2020-diverse} and IR-based utility \citep{boudin-gallina-2021-redefining} reflect major efforts to move beyond reference-based evaluation. Building upon these works, \textsc{KPEval} aims to provide a unified perspective and to advance the evaluation methodology. Figure \ref{main-framework} illustrates the evaluation design for each aspect, which we will introduce next.

\setlength{\tabcolsep}{3pt}
\begin{table}
    \centering
    \resizebox{\linewidth}{!}{%
    \begin{tabular}{c | c c c c}
    \hline
     & \texttt{KP-Set} & \texttt{Input} & \texttt{Reference} & \texttt{Corpus}  \\
    \hline
    \textbf{{Reference Agreement}} & \cmark &  & \cmark &  \\
    \textbf{{Faithfulness}} &  & \cmark &  &  \\
    \textbf{{Diversity}} & \cmark &  &  &  \\
    \textbf{{Utility}} & \cmark & \cmark &  & \cmark \\
    \hline
    \end{tabular}
    }
    \vspace{-2mm}
    \caption{Assumptions of \textsc{KPEval}'s aspects: whether they operate on a set of keyphrases (\texttt{KP-Set}) and whether they require input, reference, or a corpus.}
    \label{tab:kp-sys-property-assumptions}
    \vspace{-4mm}
\end{table}

\section{\textsc{KPEval}: Reference-Based Evaluation with Semantic Matching}
\label{ref-based-eval}

To begin with, we focus on reference agreement, the most extensively investigated aspect. Recognizing the limitations of previous approaches, we introduce a semantic matching formulation and conduct meta-evaluation to confirm its effectiveness.

\subsection{Reference Agreement: Metric Design}
\label{saliency-metric-design}
\fcolorbox{teal!30}{green!10}{\parbox{0.465\textwidth}{
\textit{\textbf{Desiderata}: a prediction should be credited if it is semantically similar to a human-written keyphrase; matching should be at phrase-level.}}
}
\vspace{0.3pt}

Despite the prevalent use of existing reference-based metrics, their designs harbor intrinsic limitations. On one hand, $F1@5$ \citep{meng-etal-2017-deep} and $F1@M$ \citep{yuan-etal-2020-one} fail to credit many semantically correct predictions. On the other hand, BertScore with concatenated predictions and references \citep{koto-etal-2022-lipkey} reflects semantic similarity, but its token-level matching strategy obscures the semantics of individual keyphrases. Recognizing these limitations, we propose a \textit{phrase-level semantic matching} strategy in \textsc{KPEval} and define semantic precision ($SemP$), recall ($SemR$), and F1 ($SemF1$) as follows\footnote{Out metric should be distinguished from \citet{bansal-etal-2022-sem}. We keep the name choice as the tasks are different.}:
\begin{equation}
\resizebox{0.8\hsize}{!}{$SemP(\mathcal{P},\mathcal{Y})=\frac{1}{|\mathcal{P}|}\sum_{p\in\mathcal{P}}\max_{y\in\mathcal{Y}}sim(p,y)$}\nonumber,
\end{equation}
\begin{equation}
\resizebox{0.8\hsize}{!}{$SemR(\mathcal{P},\mathcal{Y})=\frac{1}{|\mathcal{Y}|}\sum_{y\in\mathcal{Y}}\max_{p\in\mathcal{P}}sim(p,y)$}\nonumber,
\end{equation}
\begin{equation}
\resizebox{0.75\hsize}{!}{$SemF1(\mathcal{P},\mathcal{Y})=\frac{2\cdot SemP(\mathcal{P},\mathcal{Y})\cdot SemR(\mathcal{P},\mathcal{Y})}{SemP(\mathcal{P},\mathcal{Y}) + SemR(\mathcal{P},\mathcal{Y})}$}\nonumber,
\end{equation}
where $sim$ is the similarity between the representation of two phrases. To enable the use of any existing dense embedding model, in this paper, we operationalize $sim$ with the cosine similarity:
\begin{equation}
\resizebox{0.8\hsize}{!}{$sim(p,q)=cos\_sim(h_p,h_q)=\frac{h_p^Th_q}{||h_p||\cdot||h_q||}$}\nonumber,
\end{equation}
where $h_p$ is the representation of phrase $p$ obtained by aggregating the representation of all tokens in the phrase. To obtain a high quality embedding that captures phrase-level semantics well, we fine-tune a paraphrase model from \citet{reimers-gurevych-2019-sentence}\footnote{We use the checkpoint at \url{https://huggingface.co/sentence-transformers/all-mpnet-base-v2}.} using unsupervised SimCSE \citep{gao-etal-2021-simcse} on 1.04 million keyphrases from the training sets of KP20k, KPTimes, StackEx \citep{yuan-etal-2020-one}, and OpenKP \citep{xiong-etal-2019-open}. The data covers a wide range of domains including science, news, forum, and web documents. At inference time, a single phrase $p$ is provided as the input to the model, and the last hidden states are mean-pooled to obtain $h_p$. We further document the training details of this model in Appendix \ref{phrase-embedding-details}.

\subsection{Meta-Evaluation Setup}
\label{meta-evaluation-setup}
We conduct rigorous meta-evaluation to compare $SemF1$ with existing metrics. We sample 50 documents from the test sets of KP20k and KPTimes each. For each document, we obtain predictions from five representative models:  MultipartiteRank, CatSeq, SetTrans, in-domain BART models from \citet{wu-etal-2023-rethinking}, as well as five-shot prompting GPT-3.5\footnote{We use SciBART+OAGKX for KP20k and KeyBART for KPTimes to represent in-domain BART models. Detail regarding the evaluated models are provided in Appendix \ref{kp-system-impl}.}. This variety encompasses both KPE and KPG model, and includes unsupervised, supervised, and few-shot prompting methods. Then, three crowd-source annotators are asked to rate on Likert-scale the \textit{semantic similarity} between (1) each prediction keyphrase $p_i$ and the most semantically similar keyphrase in $\mathcal{Y}$, and (2) each reference keyphrase $y_i$ and the most semantically similar keyphrase in $\mathcal{P}$. We report the details of the annotator recruitment process, the annotation instructions, and the interface in Appendix \ref{human-eval-setup}. 

A total of 1500 document-level annotations with 13401 phrase-level evaluations are collected. As presented in Table \ref{tab:iaa}, we observe 0.75 Krippendorff's alpha for both datasets and both matching directions, indicating a high inter-annotator agreement. The annotations are aggregated to obtain (1) \textit{phrase-level} scores for matching a single phrase to a set of phrases ($p_i\rightarrow\mathcal{Y}$ and $y_i\rightarrow\mathcal{P}$) and (2) \textit{document-level} precision, recall, and F1 scores, calculated after normalizing the scores to a 0-1 scale. 

\subsection{Meta-Evaluation Results}
\label{ref-based-eval-meta-eval}
\label{meta-evaluation}

\setlength{\tabcolsep}{3.5pt}
\begin{table}[t]
    \centering
    \resizebox{\linewidth}{!}{
    \begin{tabular}{l | c c c | c c c }
    \hline
    & \multicolumn{3}{c|}{\textbf{Reference Agreement}} & \multicolumn{3}{c}{\textbf{Faithfulness}}\\
    & $p_i\rightarrow\mathcal{Y}$ & $y_i\rightarrow\mathcal{P}$ & average & PKP & AKP & all\\
    \hline
    \textbf{KP20k} & 0.735 & 0.750 & 0.743 & 0.644 & 0.526 & 0.627 \\
    \textbf{KPTimes} & 0.788 & 0.741 & 0.765 & 0.606 & 0.514 & 0.563 \\
    \hline
    \end{tabular}
    }
    \vspace{-2mm}
    \caption{Inter-annotator agreement measured via the interval Krippendorff's alpha. $\rightarrow$ denotes the direction of matching a single phrase against a set of phrases. "PKP" and "AKP" denote present and absent keyphrases.}
    \label{tab:iaa}
    \vspace{-4mm}
    
\end{table}

Using the \textit{document-level} F1 score annotations, we compare $SemF1$ with six baseline metrics: 

\begin{compactenum}
    \item Exact Matching $F1@M$ \citep{yuan-etal-2020-one}.
    \item $F1@M$ with Substring Matching. We conclude a match between two phrases if either one is a substring of the other. This corresponds to the INCLUDES and PARTOF strategy in \citet{zesch-gurevych-2009-approximate}.
    \item R-precision \citep{zesch-gurevych-2009-approximate}.
    \item \textit{FG} \citep{luo-etal-2021-keyphrase-generation}.
    \item Rouge-L $F1$ \citep{lin-2004-rouge}.
    \item BertScore $F Score$ \citep{bert-score}\footnote{We use the RoBERTa-large model and the representation at the 17\textsuperscript{th} layer, as recommended by the official implementation at \url{https://github.com/Tiiiger/bert_score}.}. We concatenate all the phrases in $\mathcal{P}$ with commas to form a single prediction string, and do the same for $\mathcal{Y}$ to form the reference string\footnote{We find that BertScore is insensitive to the order of labels and predictions. Details are discussed in Appendix \ref{bertscore-order-sensitivity}.}. 
\end{compactenum}

We apply Porter Stemmer \citep{Porter1980AnAF} on $\mathcal{P}$ and  $\mathcal{Y}$ before calculating baseline 1, 2, and 3.

\begin{figure}
\includegraphics[width=\columnwidth]{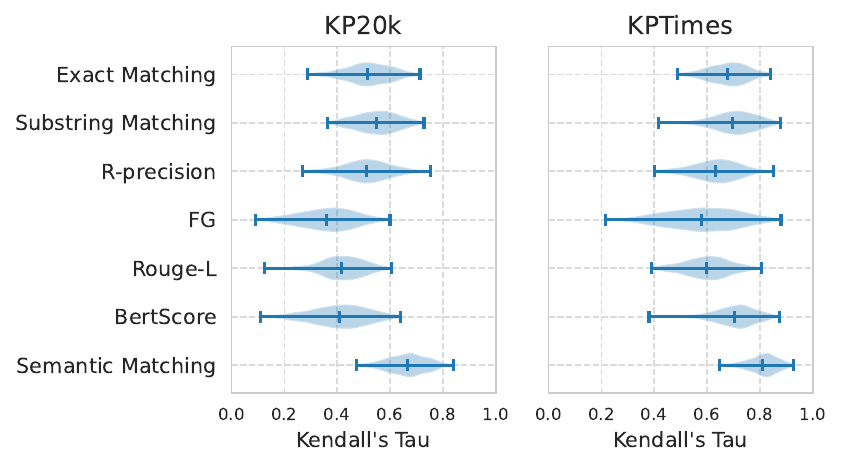}
\vspace{-8mm}
\caption{The 95\% confidence intervals for the Kendall's Tau between human and automatic metrics on KP20k and KPTimes. $SemF1$ exhibits a higher correlation with humans and smaller intervals.}
\vspace{-2mm}
\label{document-level-tau-intervals}
\end{figure}

In Figure \ref{document-level-tau-intervals}, we report the 95\% confidence interval of Kendall's Tau via input-level bootstrap resampling with 1000 samples, following \citet{deutsch-etal-2021-statistical}. Surprisingly, although exact matching produces many false negatives, \textit{existing proposals to relax exact matching do not provide much overall performance gains} either: while substring matching consistently outperforms exact matching by a small amount, R-precision and FG have a lower correlation with human compared to exact matching. BertScore's performance is highly domain-dependent: it achieves the second-best performance on KPTimes while performs poorly on KP20k. By contrast, $SemF1$ greatly outperforms other metrics on both datasets, with a much higher mean score and a smaller variation. 

  
\setlength{\tabcolsep}{3.5pt}
\begin{table}
  \centering
  \resizebox{\linewidth}{!}{
  \begin{tabular}{l | c c c | c c c  }
    \hline
    &   \multicolumn{3}{c|}{\textbf{KP20k}} & \multicolumn{3}{c}{\textbf{KPTimes}} \\
    & $r$ & $\rho$ & $\tau$ & $r$ & $\rho$ & $\tau$ \\
    \hline
    \textit{Exact Matching F1@M} & 0.705 & 0.689 & 0.590 & 0.768 & 0.792 & 0.684 \\
    \hdashline
    FastText & 0.808 & 0.797 & 0.635 & 0.844 & 0.850 & 0.693 \\
    Phrase-BERT & 0.792 & 0.793 & 0.631 & 0.876 & 0.898 & 0.755  \\
    SpanBERT & 0.757 & 0.724 & 0.562 & 0.824 & 0.819 & 0.653  \\
    SimCSE (unsupervised) & 0.862 & 0.834 & 0.677 & 0.906 & 0.905 & 0.762 \\
    SimCSE (supervised) & 0.844 & 0.827 & 0.669 & 0.907 & 0.915 & 0.780 \\
    Ours & \textbf{0.898} & \textbf{0.884} & \textbf{0.735} & \textbf{0.926} & \textbf{0.925} & \textbf{0.794} \\
    \ \ \ \ - fine-tuning & 0.870 & 0.856 & 0.705 & 0.911 & 0.915 & 0.780  \\
  \hline
  \end{tabular}
  }
  \caption{A comparison between different phrase embedding models on scoring an individual phrase against a set of phrases. Our model achieves the best matching correlation with humans, significantly outperforming the second best with $p < 0.001$ via a paired t-test.}
  \label{tab:meta-eval-embedding}
  
\end{table}

  

Is the observed high performance of $SemF1$ consistent with any embedding model? Our ablation studies suggest a negative answer. We evaluate with fine-grained \textit{phrase-level annotations} for both directions of matching (i.e., $p_i\rightarrow\mathcal{Y}$ and $y_i\rightarrow\mathcal{P}$). Table \ref{tab:meta-eval-embedding} presents the Pearson correlation ($r$), Spearman correlation ($\rho$), and Kendall's Tau ($\tau$) of exact matching and semantic matching with various embedding models: FastText \citep{joulin2016fasttext}\footnote{We use the \texttt{crawl-300d-2M.vec} model.}, Phrase-BERT \citep{wang-etal-2021-phrase}, SpanBERT \citep{joshi-etal-2020-spanbert}, SimCSE \citep{gao-etal-2021-simcse}\footnote{We use the \texttt{simcse-roberta-large} models distributed by the original authors on huggingface.}, our phrase embedding, and the model before the proposed fine-tuning. Despite being a strong strategy by design, semantic matching fails to outperform exact matching in Kendall's Tau with SpanBERT embedding. With the proposed model, semantic matching outperforms exact matching by 0.1 absolute points in Kendall's Tau and more than 0.15 absolute points in Pearson and Spearman correlation. It is worth-noting that although the base SBERT model already achieves strong performance, our phrase-level contrastive fine-tuning provides further performance gains.

\paragraph{Remark} We have confirmed that the semantic matching strategy better accommodates semantically correct \textit{predictions}. Additionally, our preliminary study indicates that human references often exhibit lexical variability. When faced with such variability, $SemF1$ demonstrates lower variance than $F1@M$ (detailed in Appendix \ref{label-variation}). 

\section{\textsc{KPEval}: Reference-Free Evaluation}
\label{ref-free-eval}

For a range of text generation tasks, the optimal output is often highly aspect-specific \citep{wen-etal-2015-semantically,mehri-eskenazi-2020-usr,fabbri-etal-2021-summeval}. As such, reference-based evaluation is incomplete as it does not always align with the evaluation goals. To address this gap, \textsc{KPEval} introduces three novel evaluation aspects, along with corresponding reference-free metrics, aimed at aligning closer with real-world application requirements.

\subsection{Faithfulness}
\label{nat-faith-methods}
\fcolorbox{teal!30}{green!10}{\parbox{0.465\textwidth}{
\textit{\textbf{Desiderata}: keyphrase predictions should always be grounded in the document.}}
}
\vspace{0.3pt}

In practical scenarios, it is vital for keyphrase systems to refrain from producing concepts not covered in the document, which we term as unfaithful keyphrases. Determining whether a keyphrase is faithful is non-trivial: an absent keyphrase could be faithful by being synonyms or parent/child concepts of the concepts in the document, while a present keyphrase could be deemed unfaithful if it has a wrong boundary. For instance, the keyphrase "hard problem" is unfaithful to a document discussing "NP-hard problem". This example also illustrates the inadequacy of reference-based evaluation, as "hard problem" may achieve a high score when matched against "NP-hard problem". 

\setlength{\tabcolsep}{3.5pt}
\begin{table}[]
  \centering
  \resizebox{\columnwidth}{!}{%
  \begin{tabular}{l | c | c | c }
    \hline
    \textbf{ID} &  \textbf{Model} & \textbf{KP20k} & \textbf{KPTimes} \\
    \hline
    M4 & MultipartiteRank & 0.694 & 0.829 \\
    M10 & CatSeq & 0.722 & 0.936 \\
    M15 & SetTrans & 0.777 & 0.912 \\
    M18 & KeyBART & - & 0.933 \\
    M19 & SciBART-large+OAGKX & 0.779 & - \\
    M21 & text-davinci-003 (5-shot) & 0.750 & 0.966 \\
  \hline
  \end{tabular}
  }
  \caption{Faithfulness evaluation of the predictions made by five models. We report the proportion of keyphrases marked as faithful by human annotators. }
  \label{tab:model-faithfulness}
  
\end{table}

\paragraph{Are existing keyphrase models faithful?} We conduct a human evaluation of the same set of five models in \cref{meta-evaluation-setup} on 100 documents from KP20k and KPTimes each. For each (document, keyphrase prediction) pair, three annotators are asked to make a binary judgement between faithful and unfaithful (details in Appendix \ref{human-eval-setup}). Table \ref{tab:iaa} presents the inter-annotator agreement. We find a moderate agreement for present keyphrases and a lower agreement for absent keyphrases. We aggregate the present keyphrase annotations by majority voting. For the absent keyphrases, two of the authors manually resolve the instances where the crowd-source annotators do not agree. Table \ref{tab:model-faithfulness} presents the faithfulness scores for the evaluated models. Surprisingly, M3's outputs are not as faithful as the neural KPG models, supporting the hypothesis that extractive models can suffer from the boundary mistakes that harm their faithfulness. In addition, models make more unfaithful predictions in KP20k compared to KPTimes, indicating the possible difficulty of accurately generating concepts grounded in scientific papers compared to news documents. 

\paragraph{Automatic Faithfulness Evaluation} Using the human annotations, we evaluate three automatic metrics for judging a keyphrase's faithfulness:
\begin{compactenum}
    \item The precision metric of \textbf{BertScore}($\mathcal{X}$, $p_i$). We use the RoBERTa-large model as in \cref{meta-evaluation}.
    \item The faithfulness metric of \textbf{BartScore} \citep{bart-score}. $p_i$ is embedded into \textcolor{codegreen}{"\textit{In summary, this is a document about }$p_i$"} for calculating its probability given $\mathcal{X}$. BART-large trained on CNN-DM \citep{see-etal-2017-get} is used.
    \item The consistency metric of \textbf{UniEval} \citep{zhong-etal-2022-towards}, which scores text generation as boolean QA. We embed $\mathcal{X}$ and $p_i$ with a template for summarization evaluation: \textcolor{codegreen}{"\textit{question: Is this claim consistent with the document? </s> summary:} \textit{the document discusses about} $p_i$\textit{. </s> document:} $\mathcal{X}$"}. Then, we use the UniEval model for summarization evaluation provided by the original authors to obtain a score expressed as the probability of the model generating "Yes" normalized by the probability for "Yes" and "No".  

\end{compactenum}

All of these metrics output a real number score. To compare their performance, we report their AUROC in Table \ref{tab:meta-eval-faithfulness}. On both datasets, UniEval outperforms BertScore and BartScore, achieving the highest agreement with human raters. Currently, \textsc{KPEval} adopts UniEval as the default faithfulness metric. We encourage future work to continue developing stronger metrics for this aspect.

  

\setlength{\tabcolsep}{3.5pt}
\begin{table}[]
  \centering
  \resizebox{0.57\linewidth}{!}{
  \begin{tabular}{l | c c }
    \hline
    & KP20k & KPTimes \\
    \hline
    \textbf{BertScore} & 0.676 & 0.648 \\
    \textbf{BartScore} & 0.677 & 0.663 \\
    \textbf{UniEval} &\textbf{0.690}$^\dagger$ & \textbf{0.672}$^\dagger$ \\
  \hline
  \end{tabular}
  }
  \caption{Meta-evaluation results for faithfulness metrics. We report the AUROC evaluated with human annotations in Table \ref{tab:model-faithfulness}. $^\dagger$statistically significantly better than the second best with $p < 0.01$ via a paired t-test.}
  \label{tab:meta-eval-faithfulness}
  
\end{table}

\subsection{Diversity}
\label{diversity-def}
\fcolorbox{teal!30}{green!10}{\parbox{0.465\textwidth}{
\textit{\textbf{Desiderata}: reward more semantically distinct concepts and penalize repetitions.}}
}
\vspace{0.3pt}

Generating keyphrases with minimal repetition is a desirable property of keyphrase applications. To assess the diversity of $\mathcal{P}$, \textsc{KPEval} includes one lexical and one semantic metric based on \citet{bahuleyan-el-asri-2020-diverse}. The lexical metric $dup\_token\_ratio$ is the percentage of duplicate tokens after stemming. The semantic metric $dup\_emb\_sim$ is the average of pairwise cosine similarity, using the phrase embedding in \cref{saliency-metric-design}:
\begin{equation}
\resizebox{0.8\hsize}{!}{$emb\_sim(\mathcal{P})=\frac{\sum_{i=1}^m\sum_{j=1}^m\mathbbm{1}(i\ne j)sim(p_i,p_j)}{m(m-1)}$}\nonumber.
\end{equation}
We note that by design, we do not penalize over-generating uninformative keyphrases, as it intuitively implies a high diversity\footnote{As a result, metrics such as the orthogonal regularization term used by CatSeqD \citep{yuan-etal-2020-one} are not suitable for our purposes, as the term naturally increases with $|\mathcal{P}|$.}. Judging the quality of the keyphrases is instead delegated to the metrics for reference agreement and faithfulness.

\subsection{Utility}
\label{utility-def}

\fcolorbox{teal!30}{green!10}{\parbox{0.465\textwidth}{
\textit{\textbf{Desiderata}: reward predictions that facilitate effective ad-hoc retrieval of the document.}}
}
\vspace{0.3pt}

Information Retrieval (IR) is an important downstream application for keyphrases \citep{10.1145/312624.312671, 10.1145/1141753.1141800, kim-etal-2013-applying}. To directly evaluate whether $\mathcal{M}$ can generate useful keyphrases for IR-related tasks, \textsc{KPEval} tests $\mathcal{P}$ on facilitating \textit{ad-hoc retrieval} of $\mathcal{X}$ from an in-domain corpus $\mathcal{C}$ \citep{boudin-gallina-2021-redefining}. 

Concretely, we leverage an in-domain corpus $\mathcal{C}$ that has documents and human-annotated keyphrases\footnote{We use the respective training sets as $\mathcal{C}$ for KP20k and KPTimes. In practice, one can run any keyphrase prediction method if human-written keyphrases are not available for $\mathcal{C}$.}. We first index $\mathcal{C}$'s documents into the form \textcolor{codegreen}{\textit{(title, keyphrases)} $\rightarrow$ \textit{document}}. To evaluate $\mathcal{P}$, we add a single entry \textcolor{codegreen}{\textit{($\mathcal{X}$'s title, $\mathcal{P}$)} $\rightarrow \mathcal{X}$} to the aforementioned database. Then, a set of queries $\mathcal{Q}=\{q_1,q_2,...,q_{|Q|}\}$ specifically written for $\mathcal{X}$ are used to attempt to retrieve $\mathcal{X}$ from this pool. The utility of $\mathcal{P}$ is measured by two metrics for retrieval effectiveness: Recall at $k$ ($Recall@k$) and Reciprocal Rank at $k$ ($RR@k$), averaged across all queries in $\mathcal{Q}$. To simulate the queries written by real users, we use GPT-4 \citep{OpenAI2023GPT4TR} to annotate three ad-hoc queries based on each document. For KP20k, the queries are in the style of in-text citations similar to \citet{boudin-gallina-2021-redefining}. For KPTimes, we generate short phrases that mimic queries on web search engines. We present the prompting details in Appendix \ref{adhoc-query-construction}. For metric calculation, we consider BM25 \citep{10.1007/978-1-4471-2099-5_24} and a dense embedding model\footnote{We use \texttt{cross-encoder/ms-marco-MiniLM-L-6-v2} model via huggingface.} as the retriever and report the averaged scores.

\section{Fine-grained benchmarking of keyphrase systems}
\label{re-eval}

Finally, we benchmark 23 keyphrase systems with \textsc{KPEval}, with the full evaluation results on KP20k and KPTimes presented in Table \ref{tab:all-results-main}. The implementation details are presented in Appendix \ref{kp-system-impl}. This section presents the insights on two questions:
\begin{compactenum}
    \item Do our finding align with the conclusions drawn from previous model comparisons?
    \item How do large language models compare with existing keyphrase prediction methods?
\end{compactenum}

\setlength{\tabcolsep}{3.5pt}
\begin{table*}[]
 \centering
 \vspace{-3mm}
 \resizebox{\linewidth}{!}{
 \begin{tabular}{l | c | l l l  | c | l l  | l l }
 \hline
 \multirow{2}{*}{\textbf{Model}} & \multirow{2}{*}{\textbf{\#KP}} & \multicolumn{3}{c|}{\textbf{Reference Agreement ($\uparrow$)}} & \textbf{Faithfulness ($\uparrow$)} & \multicolumn{2}{c|}{\textbf{Diversity ($\downarrow$)}} & \multicolumn{2}{c}{\textbf{Utility ($\uparrow$)}} \\
& & $F1@M (P)$ & $F1@M (A)$ & $SemF1$ & $UniEval$ & $dup$ &$emb\_sim$ & $RR@5$ & $Recall@5$ \\
 \hline
(\textbf{A}) CatSeq & 7.2 & 0.362 & 0.003 & 0.548 & 0.774\textsuperscript{\textbf{B}}\textsuperscript{\textbf{D}} & 0.370 & 0.310 & 0.884\textsuperscript{\textbf{C}} & 0.797\textsuperscript{\textbf{C}} \\
(\textbf{B}) CatSeqTG-2RF1 & 7.9 & 0.386\textsuperscript{\textbf{A}}\textsuperscript{\textbf{C}} & 0.005\textsuperscript{\textbf{A}} & 0.553\textsuperscript{\textbf{A}} & 0.734 & 0.355\textsuperscript{\textbf{A}} & 0.251\textsuperscript{\textbf{A}} & 0.886\textsuperscript{\textbf{C}} & 0.801\textsuperscript{\textbf{A}}\textsuperscript{\textbf{C}} \\
(\textbf{C}) ExHiRD-h & 5.5 & 0.374\textsuperscript{\textbf{A}} & 0.005\textsuperscript{\textbf{A}} & 0.562\textsuperscript{\textbf{A}}\textsuperscript{\textbf{B}} & 0.781\textsuperscript{\textbf{A}}\textsuperscript{\textbf{B}}\textsuperscript{\textbf{D}} & 0.214\textsuperscript{\textbf{A}}\textsuperscript{\textbf{B}}\textsuperscript{\textbf{D}} & 0.195\textsuperscript{\textbf{A}}\textsuperscript{\textbf{B}} & 0.878 & 0.791 \\
(\textbf{D}) SetTrans & 7.7 & 0.390\textsuperscript{\textbf{A}}\textsuperscript{\textbf{C}} & 0.006\textsuperscript{\textbf{A}}\textsuperscript{\textbf{B}}\textsuperscript{\textbf{C}} & 0.583\textsuperscript{\textbf{A}}\textsuperscript{\textbf{B}}\textsuperscript{\textbf{C}} & 0.766\textsuperscript{\textbf{B}} & 0.308\textsuperscript{\textbf{A}}\textsuperscript{\textbf{B}} & 0.203\textsuperscript{\textbf{A}}\textsuperscript{\textbf{B}} & 0.889\textsuperscript{\textbf{A}}\textsuperscript{\textbf{C}} & 0.803\textsuperscript{\textbf{A}}\textsuperscript{\textbf{C}} \\
 \hline
 \end{tabular}
 }
 \caption{Re-evaluation of four models considered in \citet{ye-etal-2021-one2set}. \#KP = Number of keyphrases. $(P)$ and $(A)$ indicate present and absent keyphrases. $dup$ = $dup\_token\_ratio$. For $dup$ and $emb\_sim$, lower scores are better. We use four superscripts to mark the results that are significantly better than CatSeq (\textbf{A}), CatSeqTG-2RF1 (\textbf{B}), ExHiRD-h (\textbf{C}), and SetTrans (\textbf{D}) with $p<0.01$ via a paired t-test.}
 \label{tab:settrans-reeval}
\end{table*}

\begin{figure*}[t!]
\includegraphics[width=\linewidth]{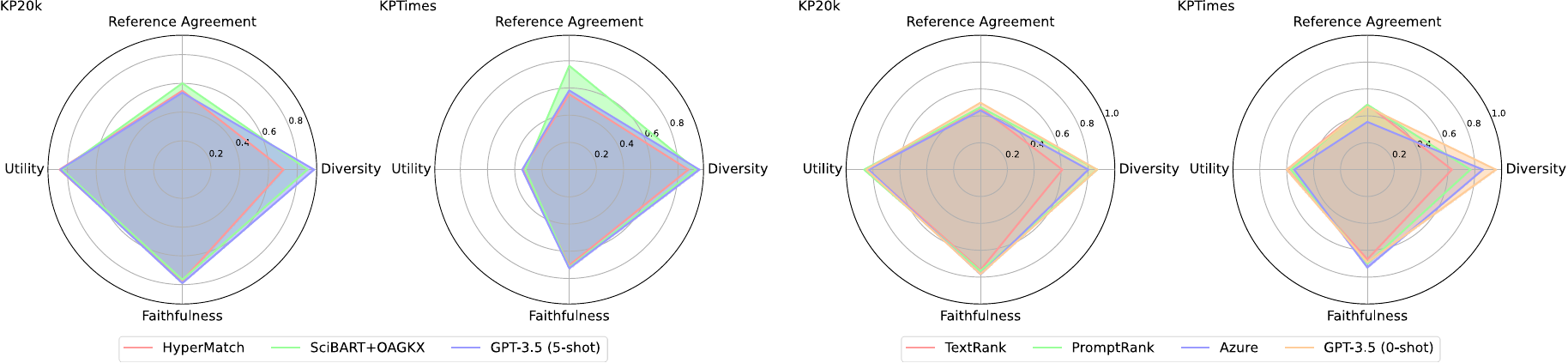}
\caption{A comparison between GPT-3.5 and strong supervised (left) and unsupervised (right) keyphrase extraction and keyphrase generation methods. GPT-3.5 achieves strong performance on both datasets and most dimensions. We use $RR@5$ to represent utility and $1-dup\_token\_ratio$ to represent diversity. }
\label{fig-radar-chart}
\end{figure*}

\paragraph{Uncovering blind-spots of established results} We revisit a set of models compared in \citet{ye-etal-2021-one2set}: CatSeq, CatSeqTG-2RF1, ExHiRD-h, and SetTrans. As shown in Table \ref{tab:settrans-reeval}, a nuanced pattern emerges when evaluating beyond $F1@M$, the main metric reported in the original paper. $SemF1$ is consistent with $F1@M$ in recognizing SetTrans as the best model for reference agreement. However, SetTrans does not outperform all the three baselines in reference-free evaluation. Specifically, the best faithfulness and diversity scores are achieved by ExHiRD-h, and the difference between SetTrans and CatSeqTG-2RF1 in utility is insignificant. Moreover, contradicting with $F1@M$, \textsc{KPEval}'s metrics show a superiority of ExHiRD-h over CatSeqTG-2RF1 for reference agreement, faithfulness, and diversity. We provide several supporting examples in Appendix \ref{qualitative-study}. By revealing these blind-spots in previous results, \textsc{KPEval} enables a holistic view in model comparison and a stricter criterion in establishing the state-of-the-art.

\paragraph{LLM vs traditional keyphrase models} With the ascent of LLMs as foundational elements in NLP, their efficacy in keyphrase prediction warrants examination. Prior research reported significant performance gaps when evaluating LLMs with exact matching $F1@M$ \citep{song2023chatgpt, martinez2023chatgpt}. With \textsc{KPEval}, we conduct a more comprehensive investigation. In Figure \ref{fig-radar-chart}, we compare GPT-3.5 with state-of-the-art KPE and KPG methods. For supervised methods, performance of five-shot prompting GPT-3.5 is comparable or better than HyperMatch along every dimension, and comparable to SciBART+OAGKX in diversity, utility, and faithfulness. In addition, zero-shot prompting outperforms the unsupervised TextRank, PromptRank, and the Azure API in diversity while being competitive across other dimensions. These results suggest that the potential of LLMs for keyphrase prediction may be underappreciated under traditional evaluation paradigms.

\paragraph{Discussion} What have we learned in this re-evaluation? First, the refined evaluation facilitated by \textsc{KPEval} challenges some of the existing model comparisons and emphasizes the difficulty of outperforming baselines across all aspects. In fact, these aspects test unique abilities, exhibiting weak cross-aspect correlations (Appendix \ref{inter-metric-correlation}) and distinct preferences for keyphrase systems (Table \ref{tab:settrans-reeval}, \ref{tab:all-results-main}). Our findings advocate for a tailored approach to metric weighting, allowing users to customize evaluations based on their evaluation desiderata. Finally, our results reveal strong performance of GPT-3.5, encouraging future work to further understand and improve LLMs for keyphrase prediction.

\setlength{\tabcolsep}{3.5pt}
\begin{table*}[!ht]
 \centering
 \vspace{-3mm}
 \resizebox{\linewidth}{!}{
 \begin{tabular}{c | c | c | c c c | c | c c | c c }
 \hline
 \multirow{2}{*}{\textbf{ID}} & \multirow{2}{*}{\textbf{Model}} & \multirow{2}{*}{\textbf{\#KP}} & \multicolumn{3}{c|}{\textbf{Reference Agreement}} & \textbf{Faithfulness} & \multicolumn{2}{c|}{\textbf{Diversity}} & \multicolumn{2}{c}{\textbf{Utility}} \\
& & & $SemP \uparrow$ & $SemR \uparrow$ & $SemF1 \uparrow$ & $UniEval\uparrow$ & $dup \downarrow$ &$emb\_sim \downarrow$ & $RR@5 \uparrow$ & $Recall@5 \uparrow$ \\
 \hline
 \multicolumn{8}{l}{\textbf{\textit{KP20k}}} \\
 \hdashline
\textbf{M1} & TF-IDF\textsuperscript{\ding{107}} & 10.0 & 0.431 & 0.524 & 0.463 & 0.696  & 0.396 & 0.161 & \underline{0.889} & \underline{0.805} \\
\textbf{M2} & TextRank\textsuperscript{\ding{107}} & 10.0 & 0.432 & 0.518 & 0.460 & 0.737  & 0.391 & 0.180 & 0.882 & 0.796 \\
\textbf{M3} & RAKE\textsuperscript{\ding{107}} & 10.0 & 0.401 & 0.499 & 0.437 & 0.733 & 0.295 & 0.170 & 0.855 & 0.783
\\
\textbf{M4} & MultipartiteRank\textsuperscript{\ding{107}} & 10.0 & 0.364 & 0.548 & 0.429 & 0.712 & 0.148 & \underline{0.092} & 0.881 & 0.794 \\
\textbf{M5} & YAKE!\textsuperscript{\ding{107}} & 10.0 & 0.434 & 0.508 & 0.459 & 0.778 & 0.372 & 0.190 & 0.859 & 0.786 \\
\textbf{M6} & PromptRank\textsuperscript{\ding{107}}\textsuperscript{\ding{95}} & 10.0 & 0.414 & 0.556 & 0.465 & 0.755 & 0.133 & 0.159 & 0.866 & 0.786 \\
\textbf{M7} & Kea\textsuperscript{\ding{107}}\textsuperscript{\ding{169}} & 10.0 & 0.452 & 0.533 & 0.479 & 0.715 & 0.445 & 0.175 & \textbf{0.892} &  \textbf{0.809}$^\dagger$ \\
\textbf{M8} & BERT+CRF\textsuperscript{\ding{107}}\textsuperscript{\ding{169}}\textsuperscript{\ding{95}} & 10.0 & \textbf{0.675}$^\dagger$ & 0.447 & 0.515 & \textbf{0.801}$^\dagger$ & 0.137 & 0.399 &  0.879 & 0.796\\
\textbf{M9} & HyperMatch\textsuperscript{\ding{107}}\textsuperscript{\ding{169}}\textsuperscript{\ding{95}} & 10.0 & 0.505 & \textbf{0.631}$^\dagger$ & 0.549 & 0.772 & 0.297 & 0.233 & 0.881 & 0.798 \\
\textbf{M10} & CatSeq\textsuperscript{\ding{169}} & 7.2 & 0.596 & 0.531 & 0.548 & 0.774 & 0.370 & 0.310 & 0.884 & 0.797 \\
\textbf{M11} & CatSeqTG-2RF1\textsuperscript{\ding{169}} & 7.9 & 0.550 & 0.579 & 0.553 & 0.734 & 0.355 & 0.251 & 0.886 & 0.801 \\
\textbf{M12} & ExHiRD-h\textsuperscript{\ding{169}} & 5.5 & 0.587 & 0.559 & 0.562 & 0.781 & 0.214 & 0.195 & 0.878 & 0.791 \\
\textbf{M13} & SEG-Net\textsuperscript{\ding{169}} & 11.0 & 0.560 & 0.596 & 0.565 & 0.757 & 0.260 & 0.177 & \underline{0.889} & 0.804 \\
\textbf{M14} & Transformer\textsuperscript{\ding{169}} & 8.2 & 0.558 & 0.593 & 0.562 & 0.747 & 0.289 & 0.239 & 0.878 & 0.791 \\
\textbf{M15} & SetTrans\textsuperscript{\ding{169}} & 7.7 & 0.570 & \underline{0.623} & \underline{0.583} & 0.766 & 0.308 & 0.203 & \underline{0.889} & 0.803 \\
\textbf{M16} & SciBERT-G\textsuperscript{\ding{169}}\textsuperscript{\ding{95}} & 6.1 & 0.587 & 0.590 & 0.577 & 0.762 & 0.152 & 0.177 & 0.878 & 0.791 \\
\textbf{M17} & BART-large\textsuperscript{\ding{169}}\textsuperscript{\ding{95}} & 6.6 & 0.567 & \textbf{0.607} & 0.574 & 0.759 & 0.144 & 0.168 & 0.877 & 0.790 \\
\textbf{M18} & KeyBART\textsuperscript{\ding{169}}\textsuperscript{\ding{95}} & 6.0 & 0.580 & 0.599 & 0.577 & 0.765 & 0.130 & 0.163 & 0.883 & 0.792 \\
\textbf{M19} & SciBART-large+OAGKX\textsuperscript{\ding{169}}\textsuperscript{\ding{95}} & 6.1 & \underline{0.601} & \underline{0.623} & \textbf{0.601}$^\dagger$ & 0.766 & \underline{0.129} & 0.158 & 0.879 & 0.793 \\
\textbf{M20} & text-davinci-003 (0-shot)\textsuperscript{\ding{95}} & 9.5 & 0.433 & 0.616 & 0.498 & 0.777 & 0.143 & 0.110 & 0.888 & 0.799 \\
\textbf{M21} & text-davinci-003 (5-shot)\textsuperscript{\ding{95}} & 6.5 & 0.503 & 0.599 & 0.535 & \underline{0.790} & \textbf{0.084}$^\dagger$ & 0.113 & 0.886 & 0.802 \\
\textbf{M22} & Amazon Comprehend\textsuperscript{\ding{107}} & 10.0 & 0.224 & 0.324 & 0.259 & 0.693 & 0.436 & 0.103 & 0.855 & 0.758 \\
\textbf{M23} & Azure Cognitive Services\textsuperscript{\ding{107}} & 10.0 & 0.387 & 0.536 & 0.440 & 0.778 & 0.198 & \textbf{0.068}$^\dagger$ & 0.874 & 0.780 \\
 \hline
 \multicolumn{8}{l}{\textbf{\textit{KPTimes}}} \\
 \hdashline
\textbf{M1} & TF-IDF\textsuperscript{\ding{107}} & 10.0 & 0.403 & 0.557 & 0.461 & 0.658 & 0.175 & 0.163 & 0.589 & 0.482 \\
\textbf{M2} & TextRank\textsuperscript{\ding{107}} & 10.0 & 0.457 & 0.532 & 0.483 & 0.667 & 0.371 & 0.246 & 0.598 & 0.485 \\
\textbf{M3} & RAKE & 10.0 & 0.284 & 0.380 & 0.321 & 0.693 & 0.157 & 0.140 & 0.555 & 0.442 \\
\textbf{M4} & MultipartiteRank\textsuperscript{\ding{107}} & 10.0 & 0.410 & 0.575 & 0.472 & 0.670 & 0.123 & 0.150 & 0.589 & 0.483 \\
\textbf{M5} & YAKE! & 10.0 & 0.387 & 0.442 & 0.405 & 0.639 & 0.317 & 0.225 & 0.564 & 0.441 \\
\textbf{M6} & PromptRank\textsuperscript{\ding{107}}\textsuperscript{\ding{95}} & 10.0 & 0.427 & 0.574 & 0.483 & 0.699 & 0.233 & \underline{0.134} & 0.569 & 0.458\\
\textbf{M7} & Kea\textsuperscript{\ding{107}}\textsuperscript{\ding{169}} & 10.0 & 0.427 & 0.575 & 0.484 & 0.699 & 0.219 & 0.175 & \textbf{0.607}$^\dagger$ & \textbf{0.500}$^\dagger$ \\
\textbf{M8} & BERT+CRF\textsuperscript{\ding{107}}\textsuperscript{\ding{169}}\textsuperscript{\ding{95}} & 2.3 & 0.741 & 0.459 & 0.550 & \textbf{0.769}$^\dagger$ & 0.072 & 0.556 & 0.583 & 0.478 \\
\textbf{M9} & HyperMatch\textsuperscript{\ding{107}}\textsuperscript{\ding{169}}\textsuperscript{\ding{95}} & 10.0 & 0.493 & 0.655 & 0.554 & 0.703 & 0.124 & 0.271 & 0.587 & 0.477 \\
\textbf{M10} & CatSeq\textsuperscript{\ding{169}} & 5.9 & 0.723 & 0.714 & 0.708 & 0.683 & 0.191 & 0.257 & 0.569 & 0.454 \\
\textbf{M11} & CatSeqTG-2RF1\textsuperscript{\ding{169}} & n/a & n/a & n/a & n/a & n/a & n/a & n/a & n/a & n/a \\
\textbf{M12} & ExHiRD-h\textsuperscript{\ding{169}} & 5.8 & 0.725 & 0.713 & 0.709 & 0.680 & 0.149 & 0.238 & 0.572 & 0.454 \\
\textbf{M13} & SEG-Net\textsuperscript{\ding{169}} & n/a & n/a & n/a & n/a & n/a & n/a & n/a & n/a & n/a \\
\textbf{M14} & Transformer\textsuperscript{\ding{169}} & 5.7 & 0.720 & 0.736 & 0.717 & 0.658 & 0.133 & 0.231 & 0.569 & 0.456 \\
\textbf{M15} & SetTrans\textsuperscript{\ding{169}} & 8.4 & 0.662 & \underline{0.801} & 0.716 & 0.645 & 0.210 & 0.232 & 0.572 & 0.454 \\
\textbf{M16} & SciBERT-G\textsuperscript{\ding{169}}\textsuperscript{\ding{95}} & 4.5 & \textbf{0.784} & 0.737 & 0.749 & 0.716 & \underline{0.064} & 0.208 & 0.579 & 0.468 \\
\textbf{M17} & BART-large\textsuperscript{\ding{169}}\textsuperscript{\ding{95}} & 5.4 & 0.768 & 0.796 & \textbf{0.770}$^\dagger$ & 0.706 & 0.074 & 0.203 & 0.582 & 0.469 \\
\textbf{M18} & KeyBART\textsuperscript{\ding{169}}\textsuperscript{\ding{95}} & 5.9 & 0.751 & \textbf{0.807}$^\dagger$ & \underline{0.766} & 0.696 & 0.086 & 0.206 &  0.579 & 0.466 \\
\textbf{M19} & SciBART-large+OAGKX\textsuperscript{\ding{169}}\textsuperscript{\ding{95}} & 4.8 & \underline{0.782} & 0.766 & 0.763 & 0.718 & 0.069 & 0.203 & 0.581 & 0.472 \\
\textbf{M20} & text-davinci-003 (0-shot)\textsuperscript{\ding{95}} & 14.1 & 0.383 & 0.620 & 0.467 & 0.690 & 0.082 & 0.137 & \underline{0.599} & \underline{0.487} \\
\textbf{M21} & text-davinci-003 (5-shot)\textsuperscript{\ding{95}} & 6.7 & 0.549 & 0.643 & 0.582 & 0.725 & \textbf{0.044}$^\dagger$ & 0.188 & 0.590 & 0.479 \\
\textbf{M22} & Amazon Comprehend\textsuperscript{\ding{107}} & 10.0 & 0.258 & 0.347 & 0.292 & 0.617 & 0.291 & 0.203 & 0.544 & 0.433 \\
\textbf{M23} & Azure Cognitive Services\textsuperscript{\ding{107}} & 10.0 & 0.312 & 0.425 & 0.355 & \underline{0.728} & 0.140 & \textbf{0.133}$^\dagger$ & 0.543 & 0.425 \\
 \hline
 \end{tabular}
 }
 \caption{Evaluation results for 23 evaluated keyphrase systems. Due to budget constraints, we only sample 1000 documents per dataset for utility evaluation. For the other aspects, the complete test sets are used. \#KP = Number of keyphrases. $dup$ = $dup\_token\_ratio$. We use $\uparrow$ for the higher the better and $\downarrow$ for the reverse. The best is boldfaced and the second best is underlined. $^\dagger$ indicates statistically significantly better than the second best with $p < 0.01$ via a paired t-test. \textsuperscript{\ding{107}} = KPE systems. \textsuperscript{\ding{169}} = supervised models. \textsuperscript{\ding{95}} = pre-trained language models. }
 \label{tab:all-results-main}
\end{table*}

\section{Conclusion}
We introduce \textsc{KPEval}, a fine-grained 
evaluation framework that conducts semantic-based evaluation on reference agreement, faithfulness, diversity, and utility of keyphrase systems. We show the advantage of our metrics via rigorous human evaluation, and exhibit the usability of \textsc{KPEval} through a large-scale evaluation of keyphrase systems including LLM-based methods and keyphrase APIs. Our framework marks the first step towards systematically evaluating keyphrase systems in the era of LLMs. We hope \textsc{KPEval} can motivate future works to adopt more accurate evaluation metrics and further advance the evaluation methodology. Future studies might also explore the development of utility metrics tailored to the specific requirements of applications in niche domains.

\section*{Limitations}
While our study sheds light on enhancing the keyphrase evaluation methodology, several limitations exist for \textsc{KPEval}. 

\begin{enumerate}
    \item \textbf{Multilingual Evaluation.} We encourage future work to extend the evaluations in this paper to multilingual setting. By design, the aspect and metric formulations in \textsc{KPEval} are language-agnostic. For instance, $SemF1$ can be implemented with multilingual embeddings. Such embeddings need not to keyphrase-specific. For instance, Table \ref{tab:meta-eval-embedding} suggests that off-the-shelf embeddings cam already have reasonable performance. 
    \item \textbf{Alternative Scoring Schemes.} \textsc{KPEval}'s evaluation and meta-evaluation strategies always target at producing fine-grained numeric scores. This is different from tasks like machine translation where direct assessment scores are annotated \citep{graham-etal-2013-continuous} or LLM competitions that report Elo scores \citep{ChatArena}. Exploring whether these schemes may provide better evaluation quality is an important question for future work on keyphrase evaluation. 
    \item \textbf{LLM-Based Evaluation.} Recent works have shown the viability of using LLMs for human-aligning aspect-specific evaluation \citep{liu-etal-2023-g}. By comprehensively establishing the possible evaluation aspects and curating meta-evaluation data for reference agreement and faithfulness, our work sets up the necessary preparations for evaluating LLM-based metrics. We encourage future work to formally define and investigate the performance of keyphrase evaluation metrics based on LLMs.
\end{enumerate}

\section*{Acknowledgments}

The research is supported in part by Taboola, NSF
CCF-2200274, and an Amazon AWS credit award.
We thank the Taboola team for helpful discussion.
We thank Amita Kamath, Wasi Ahmad, as well
as other members of the UCLA-NLP group for
providing their valuable feedback. We also thank
Jingnong Qu, Xiaoxian Shen and Xueer Li for their
help in a meta-evaluation pilot study.


\bibliography{anthology,custom}

\clearpage
\appendix
\twocolumn[{%
 \centering
 \Large\bf Supplementary Material: Appendices \\ [20pt]
}]

\section{Literature survey: evaluation methods used in recent keyphrase papers}
\label{kp-paper-eval-survey}

 We survey all the papers published from 2017 to 2023 in major conferences for AI, NLP, and IR (ACL, NAACL, EMNLP, AAAI, and SIGIR) about keyphrase extraction or keyphrase generation. We choose year 2017 as it marks the start of deep keyphrase generation methods \citep{meng-etal-2017-deep}. We manually check each paper's experiment sections and note down which of the six major categories do the reported evaluation metrics belong to: (1) precision, recall, and F1 based on exact-matching; (2) diversity metrics, such as duplication ratio; (3) ranking-based metrics such as mAP, $\alpha$-NDCG, and MRR; (4) approximate versions of exact matching such as n-gram matching; (5) retrieval-based utility metrics; (6) human evaluation. We make sure each of metrics used in the surveyed papers can fall under one and only one category under this ontology.

The survey results are presented in Figure \ref{eval-metric-choice}. Overall, despite its limitation, exact matching has been \textit{de facto} the method for assessing the performance of newly proposed keyphrase systems, and there has been limited progress in adopting alternative metrics. The majority of papers report exact matching precision, recall, and F1. Two thirds of all papers use exact matching as the only metric, including 10 out of 11 papers published in 2023. Human evaluation is only conducted in one paper surveyed \citep{bennani-smires-etal-2018-simple}. 

\section{Keyphrase Systems: Implementation Details and Full Evaluation Results}
\label{kp-system-impl}
\label{full-eval-results}

In this section, we describe in detail the considered keyphrase systems as well as how we obtain their outputs for evaluation. 

\subsection{Keyphrase Systems}

We consider three types of keyphrase systems: keyphrase extraction models, keyphrase generation models, and APIs including large language models. 

\paragraph{Keyphrase Extraction Systems}
KPE has traditionally been approached through unsupervised methods, where noun phrase candidates are ranked using heuristics \citep{hulth-2003-improved,mihalcea-tarau-2004-textrank}. Supervised approaches include feature-based ranking \citep{witten1999kea}, sequence labeling \citep{zhang-etal-2016-keyphrase}, and the use of pre-trained language models (PLMs) for task-specific objectives \citep{song-etal-2021-importance,song-etal-2022-hyperbolic}. We consider the following nine KPE models:

\begin{figure}[]
\centering
\includegraphics[width=\columnwidth]{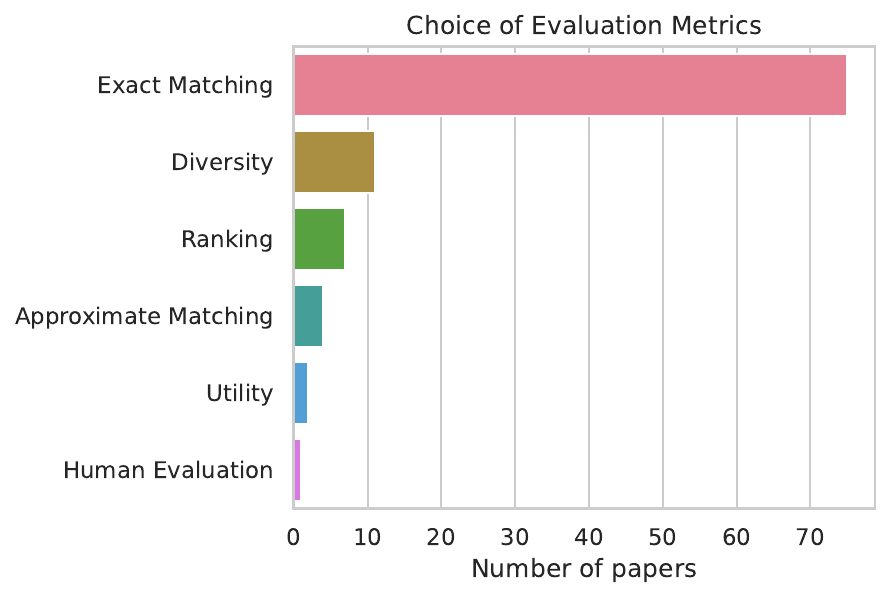}
\caption{Distribution over 75 papers of the adopted evaluation metrics: (1) F1 score based on exact-matching; (2) diversity metrics such as duplication ratio; (3) ranking-based metrics such as mAP, $\alpha$-NDCG, and MRR; (4) approximate versions of exact matching; (5) retrieval-based utility metrics; (6) human evaluation.}
\label{eval-metric-choice}
\end{figure}

\begin{compactenum}
    \item [\textbf{M1}] \textbf{TF-IDF} \citep{tf-idf} selects the phrases containing words with highest TF-IDF weight.
    \item [\textbf{M2}] \textbf{TextRank} \citep{mihalcea-tarau-2004-textrank} runs PageRank \citep{brin1998anatomy} on an undirected word cooccurrence graph. 
    \item [\textbf{M3}] \textbf{RAKE} \citep{rose2010automatic} is an efficient single-document unsupervised KPE algorithm that uses the cooccurrence graph to score keyphrase candidates.
    \item [\textbf{M4}] \textbf{MultipartiteRank} \citep{boudin-2018-unsupervised} represents the document as a multipartite graph to encode topical diversity and improve intra-topic keyphrase selection preferences.
    \item [\textbf{M5}] \textbf{YAKE!} \citep{YAKE} is an unsupervised KPE method relying on local features such as term co-occurrence and frequencies.
    \item [\textbf{M6}] \textbf{Kea} \citep{witten1999kea} builds a supervised keyphrase classifier using statistical features including TF-IDF and position information.
    \item [\textbf{M7}] \textbf{BERT+CRF} \citep{2212.10233} fine-tunes a pre-trained BERT \citep{devlin-etal-2019-bert} on sequence labeling with conditional random fields \citep{10.5555/645530.655813}. 
    \item [\textbf{M8}] \textbf{HyperMatch} \citep{song-etal-2022-hyperbolic} trains a supervised model to rank phrase-document relevance in a hyperbolic space.
    \item [\textbf{M9}] \textbf{PromptRank} \citep{kong-etal-2023-promptrank} ranks the phrases by their probability given a prompt prefix using a sequence-to-sequence pre-trained language models.
\end{compactenum}

\paragraph{Keyphrase Generation Systems}
KPG models are often trained with various supervised objectives, including \textit{One2One}, \textit{One2Seq}, and \textit{One2Set} \citep{meng-etal-2017-deep, yuan-etal-2020-one, ye-etal-2021-one2set}. A range of strategies have been proposed, including hierarchical modeling of phrases and words \citep{chen-etal-2020-exclusive}, reinforcement learning \citep{chan-etal-2019-neural}, unifying KPE and KPG \citep{chen-etal-2019-integrated,ahmad-etal-2021-select}, and using PLMs \citep{kulkarni-etal-2022-learning}. We consider ten KPG models: 

\begin{compactenum}
    \item [\textbf{M10}] \textbf{CatSeq} \citep{yuan-etal-2020-one} is an RNN trained with copy mechanism \citep{gu-etal-2016-incorporating} on generating keyphrases as a sequence.
    \item [\textbf{M11}] \textbf{CatSeqTG-2RF1} \citep{chan-etal-2019-neural} introduces an RL-based approach using recall and F1 as the rewards.
    \item [\textbf{M12}] \textbf{ExHiRD-h} \citep{chen-etal-2020-exclusive} extends CatSeq with a hierarchical decoding and an exclusion mechanism to avoid duplications.
    \item [\textbf{M13}] \textbf{SEG-Net} \citep{ahmad-etal-2021-select} unifies keyphrase extraction and keyphrase generation training and introduces layer-wise coverage attention mechanism
    \item [\textbf{M14}] \textbf{Transformer} \citep{ye-etal-2021-one2set} is a Transformer model \citep{vaswani2017attention} trained with copy mechanism on generating keyphrases as a sequence. 
    \item [\textbf{M15}] \textbf{SetTrans} \citep{ye-etal-2021-one2set} generates keyphrases in parallel based on control codes trained via a k-step target assignment process.
    \item [\textbf{M16}] \textbf{SciBERT-G} \citep{2212.10233} fine-tunes SciBERT \citep{beltagy-etal-2019-scibert} for seq2seq keyphrase generation using a prefix-LM objective \citep{arxiv.1905.03197}.
    \item [\textbf{M17}] \textbf{BART-large} \citep{2212.10233} is a BART model \citep{lewis-etal-2020-bart} fine-tuned on generating keyphrases as a sequence. 
    \item [\textbf{M18}] \textbf{KeyBART} \citep{kulkarni-etal-2022-learning} is a BART model adapted to scientific keyphrase generation before fine-tuning.
    \item [\textbf{M19}]\textbf{SciBART-large+OAGKX} \citep{2212.10233} is pre-trained on scientific corpus and scientific keyphrase generation before fine-tuning.
\end{compactenum}

\paragraph{Large language models and APIs.} Recent advancements highlight the capability of large language models (LLMs) to perform in-context learning \citep{brown2020language}. We explore GPT-3.5 \citep{ouyang2022training} for KPG in the zero-shot and few-shot prompting setting\footnote{We use OpenAI's \texttt{text-davinci-003} via API.}. We also assess two commercial keyphrase extraction APIs. 

\begin{compactenum}
    \item [\textbf{M20}] \textbf{Zero-shot prompting GPT-3.5.}
    \item [\textbf{M21}] \textbf{Five-shot prompting GPT-3.5.}
    \item [\textbf{M22}] \href{https://docs.aws.amazon.com/comprehend/latest/dg/how-key-phrases.html}{\textbf{Amazon Comprehend API}}
    \item [\textbf{M23}] \href{https://learn.microsoft.com/en-us/azure/cognitive-services/language-service/key-phrase-extraction/quickstart?pivots=programming-language-python}{\textbf{Azure Cognitive Services API}}
\end{compactenum}

\subsection{Implementation Details}

We obtain the output from the original authors for M8, M10, M11 for KP20k, and M7, M16, M17, M18 for both KP20k and KPTimes. For the other KPE and KPG models, we reproduce the results on our own. For M1, M2, M4, M6, we obtain the outputs using the \href{https://github.com/boudinfl/pke}{pke} library. For M3, M5, M8, M10, M11, M12 (KPTimes only) and M9, M13, M14, M15, we use the original implementations provided by the authors. For M18, we use the \href{https://github.com/uclanlp/DeepKPG}{DeepKPG} toolkit. For the commercial APIs, we implement the API call following the instructions. We obtained results on 3/5/2023 for M20 and 3/11/2023 for M21. Following existing KPE literature, we consider the top 10 predictions from M1, 2, 3, 4, 5, 6, 8, 9, 22, and 23. For all the systems, we truncate the input to 512 tokens. We perform hyperparameter tuning on the validation sets and ensure that the models match the performance reported by original paper or existing works such as \citet{2212.10233}. For M11 and M13 on KPTimes, we failed to obtain reasonable performance and thus choose to omit the results.

For GPT-3.5, we always start the prompt with a task definition: \textit{\textcolor{codegreen}{"Keyphrases are the phrases that summarize the most important and salient information in a document. Given a document's title and body, generate the keyphrases."}}

In the zero-shot setting, we provide the title and body in two separate lines, and start a new line with \textit{\textcolor{codegreen}{"Keyphrases (separated by comma):"}}. In the 5-shot setting, we randomly sample 5 examples from the train set for each test document, and provide their title, body, and keyphrases in the same format in the prompt before the document tested.




\section{\textsc{KPEval} Metrics: Implementation Details and Further Analyses}
\subsection{Phrase Embedding Training Details}
\label{phrase-embedding-details}

We fine-tune the paraphrase model provided by \citet{reimers-gurevych-2019-sentence} distributed at \url{https://huggingface.co/sentence-transformers/all-mpnet-base-v2}. Unsupervised SimCSE \citep{gao-etal-2021-simcse} is used as the training loss. Specifically, given a batch of $B$ phrases, the loss can be expressed as:
\begin{equation}
\resizebox{0.85\hsize}{!}{$\mathcal{L}_{simcse}=\frac{1}{B}\sum_{i=1}^B -\log\frac{e^{sim(h_i, h_i')/\tau}}{\sum_{j=1}^Be^{sim(h_i,h_j')/\tau}}$}\nonumber,
\end{equation}
where $h_i$ and $h_i'$ are the representations of phrase $i$ obtained using two separate forward passes with dropout enabled. This objective discourages the clustering of unrelated phrases in the representation space and retains a high similarity between semantically related phrase pairs. $\tau$ is a scaling factor which we empirically set to $0.05$.

We fine-tune the model on $\mathcal{L}_{simcse}$ using 1.04 million keyphrases from the training set of KP20k, KPTimes, StackEx \citep{yuan-etal-2020-one}, and OpenKP \citep{xiong-etal-2019-open}, covering a wide range of domains including science, news, forum, and web documents. We use the AdamW optimizer with maximum sequence length 12, batch size 512, dropout 0.1, and learning rate 1e-6 to fine-tune for 1 epoch. The hyperparameters are determined using a grid search on the following search space: batch size \{128, 512, 1024, 2048\}, learning rate \{1e-6, 5e-6, 1e-5, 5e-5\}. We randomly hold out 0.5\% from the training data for validation and model selection. The final training takes 30 minutes on a single Nvidia GeForce RTX 2080 Ti GPU.

\paragraph{Remark on embedding quality} In Table \ref{tab:alignment-uniformity}, we provide an additional study on the trained embedding. Specifically, following \citet{gao-etal-2021-simcse, wang2020understanding}, we evaluate alignment, the average similarity between keyphrases of similar meanings, and uniformity, the average similarity between unrelated keyphrase pairs. For alignment, we utilize the name-variation pairs constructed by \citet{chan-etal-2019-neural}. We find that our model achieves the best uniformity, which means that it assigns close to 0 similarity for unrelated pairs. For phrases with similar meanings, it achieves 0.58 alignment, which is also close to human perceptions. Finally, the separation between uniformity and alignment is also the largest for our embedding model.

\setlength{\tabcolsep}{3.5pt}
\begin{table}[]
  \centering
  \resizebox{0.8\linewidth}{!}{
  \begin{tabular}{l | c c c }
  \hline
  & Alignment & Uniformity & $\Delta$\\
  \hline
    Phrase-BERT & \textbf{0.78} & 0.54 & 0.24 \\
    SpanBERT & 0.71 & 0.54 & 0.17 \\
    Unsup. SimCSE & 0.62 & 0.23 & 0.39 \\
    Sup. SimCSE & 0.72 & 0.41 & 0.31 \\
    SBERT & 0.63 & 0.11 & 0.52 \\
    Ours & 0.58 & \textbf{0.02} & \textbf{0.56} \\
  \hline
  \end{tabular}
  }
  \caption{A comparison between different phrase embedding models. Our model achieves a large difference between alignment and uniformity, indicating the best ability to distinguish unrelated phrases.}
  \label{tab:alignment-uniformity}
  
\end{table}

\subsection{Ad-hoc Query Construction for Utility}
\label{adhoc-query-construction}

We use GPT-4 \citep{OpenAI2023GPT4TR} to annotate three ad-hoc queries per document from KP20k and KPTimes test sets. For both datasets, we sample with temperature set to 0.9 to balance quality and diversity. Due to budget constraints, we sample 1000 documents per dataset to construct the evaluation set. The prompts are presented in Figure \ref{adhoc-query-prompt}. 

\subsection{Inter-metric Correlations}
\label{inter-metric-correlation}

\begin{figure*}[]
\includegraphics[width=\textwidth]{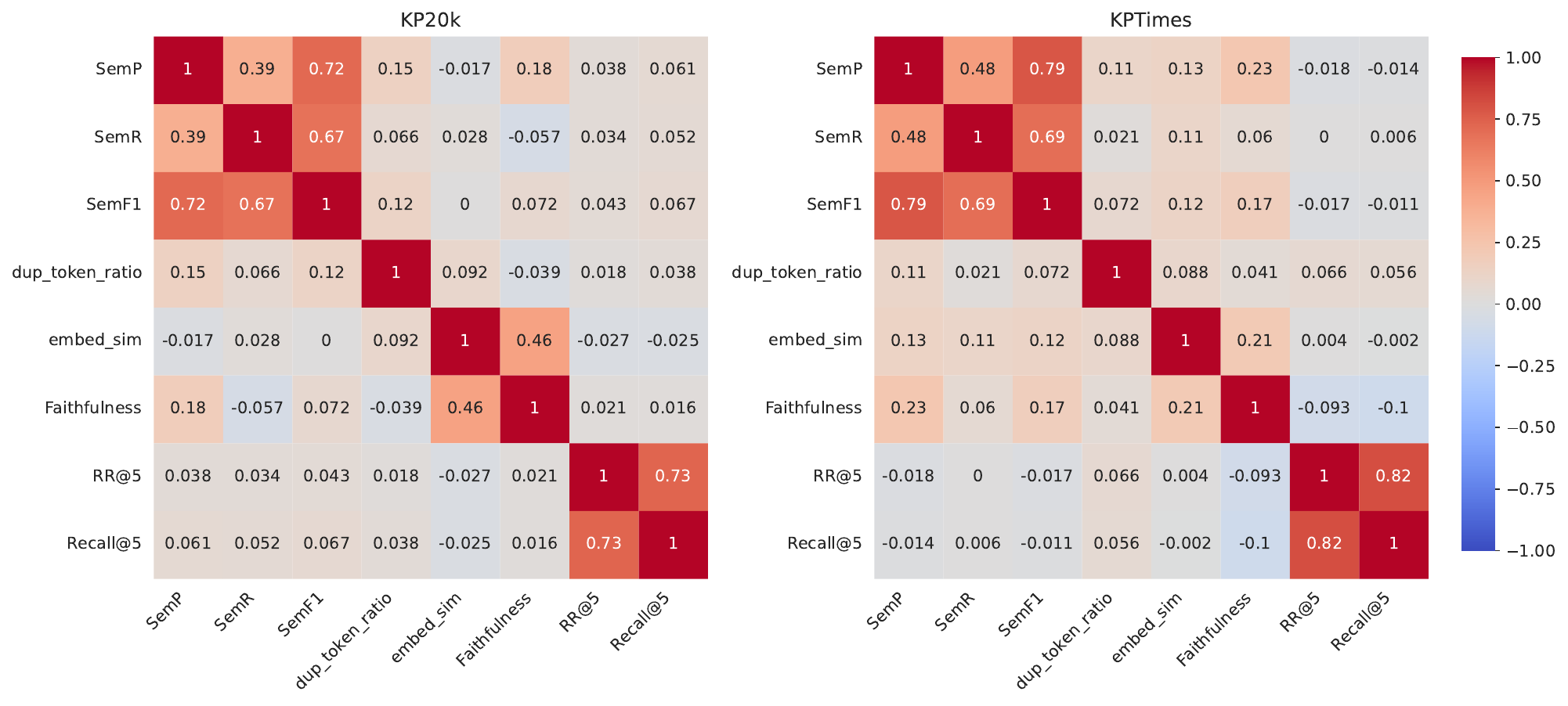}
\caption{Correlation between \textsc{KPEval}'s metrics, measured by Kendall's Tau. The diversity scores are negated to provide a more intuitive view. We find that only the metrics for the same dimension correlate with each other. }
\label{kpeval-metric-corr}
\end{figure*}

Using the document-level evaluation scores of the 21 keyphrase systems, we calculate the pair-wise Kendall's Tau for all the metrics in \textsc{KPEval}. The results are shown in Figure \ref{kpeval-metric-corr}. Overall, we find that only the metrics for the same dimension show a moderate or strong correlation with each other, and the metrics for different aspects hardly correlate. This results suggest that \textsc{KPEval}'s aspects measure distinct abilities and that optimizing a single metric does not automatically transfer to a superior performance on the other aspects. 

\subsection{Order Sensitivity of BertScore}
\label{bertscore-order-sensitivity}

\setlength{\tabcolsep}{3.5pt}
\begin{table}[]
  \centering
  \resizebox{0.9\linewidth}{!}{
  \begin{tabular}{c | c | l l l }
    \hline
    Dataset & Order & \multicolumn{1}{c}{$r$} & \multicolumn{1}{c}{$\rho$} & \multicolumn{1}{c}{$\tau$} \\
    \hline
    \multirow{2}{*}{\textbf{KP20k}} & Default & 0.562 & 0.597 & 0.436 \\
     & Permuted & 0.545$_4$ & 0.578$_5$ & 0.425$_4$ \\
     \hdashline
     \multirow{2}{*}{\textbf{KPTimes}} & Default & 0.694 & 0.729 & 0.553 \\
     & Permuted & 0.691$_3$ & 0.727$_3$ & 0.552$_3$ \\
  \hline
  \end{tabular}
  }
  \caption{The stability of BertScore when given permuted labels and references. $0.545_4$ denotes mean 0.545 with standard deviation 0.004. }
  \label{tab:bertscore-stability}
  
\end{table}

As BertScore is evaluated on two sequences instead of two sets of phrases, previous works concatenate all the predicted phrases as a single string and evaluate against all the references concatenated together \citep{koto-etal-2022-lipkey, glazkova2022applying}. However, it is unclear how to order the prediction and reference keyphrases within these two strings and whether BertScore's performance is sensitive to this ordering or not. We conduct phrase-level meta-evaluation of BertScore with phrase-level permutation applied to the matching target. Specifically, we shuffle the labels and the references before calculating meta-evaluation metrics. We repeat this process for 100 times and report the mean and standard deviation of human correlation in Table \ref{tab:bertscore-stability}. Overall, we find that the metric-human correlation of BertScore is relatively insensitive to permutations: when given reference phrases or prediction phrases concatenated in different orders, BertScore maintains a similar evaluation quality. One notable pattern is that when the references and the predictions are not permuted, BertScore obtains slightly higher performance. We hypothesize that in this case many phrases may present in the same order in the reference and the prediction, making the exactly matched instances easier to distinguish.

\subsection{Variation in Keyphrase Annotations Motivates Semantic Matching}
\label{label-variation}

A major motivation for semantic matching is that valid predictions vary in many ways. But at the same time, \textit{do human references also exhibit lexical variations}? We investigate with a pilot study of model-in-the-loop keyphrase annotation. 

\begin{figure*}[]
\centering
\fbox{\includegraphics[width=0.9\linewidth]{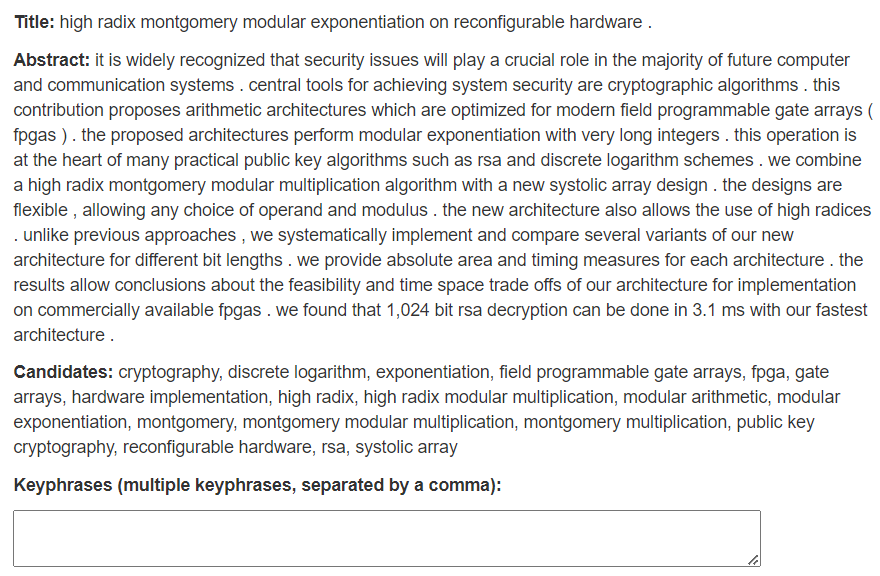}}
\caption{An example of the annotation interface for the keyphrase annotation study in Appendix \ref{label-variation}.}
\label{labeling-interface}
\end{figure*}

\paragraph{Setup}
We sample 100 documents each from the test sets of KP20k and KPTimes and combine each document's $\mathcal{Y}$ with $\mathcal{P}$ from four systems: M8, M10, M15, and M18 (KPTimes only)/M19 (KP20k only). Three MTurk annotators are presented with the document and the phrases re-ordered alphabetically. They are then asked to write keyphrases that best capture the salient information. We state that they may select from the provided keyphrases or write new keyphrases. Figure \ref{labeling-interface} presents the annotation interface. We use the same set of annotators in \cref{human-eval-setup} and collect 3226 keyphrase annotations, which approximately cost \$700.


\begin{figure}[]
\centering
\vspace{-3mm}
\includegraphics[width=0.49\textwidth]{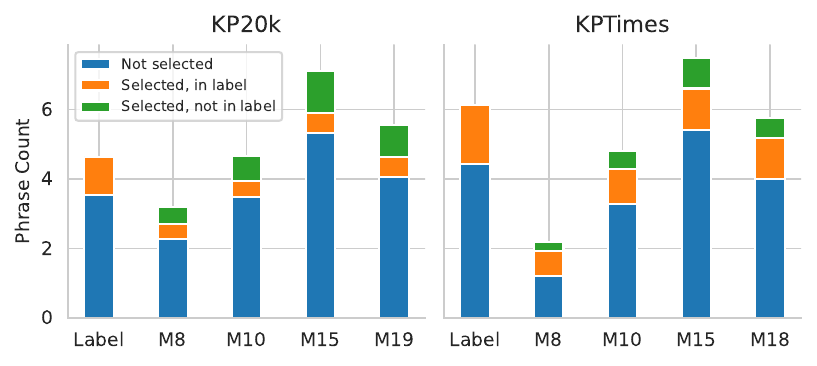}
\vspace{-8mm}
\caption{Annotators' selection distribution for each of the keyphrase sources. The reported counts are averaged across three annotators. Annotators do not prefer selecting keyphrases belonging to the original labels.}
\vspace{-4mm}
\label{selection-preference}
\end{figure}

\paragraph{Keyphrase annotations exhibit lexical variations.} Figure \ref{selection-preference} presents the distribution of phrase selected by the annotators from each source. The reported counts are averaged over three annotators. Surprisingly, we find that the keyphrases in the original labels are not preferred over the outputs from other models. First, nearly 70\% keyphrases in the original labels are not selected. Second, the annotators select several keyphrases from each model's outputs that do not appear in the label set. For KP20k, the annotators even select more such keyphrases compared to the phrases from the labels. This suggests that \textit{label variations} can be common in keyphrase annotations, even if candidate keyphrases are given as a guidance. 

Are the observed label variations caused by annotators writing entirely different concepts? We find that the average character-level edit distance between the selected phrases and the closest phrase in label keyphrases is 11.0 for KP20k and 7.0 for KPTimes, much smaller than the metric for phrases that are not selected (17.5 for KP20k and 14.6 for KPTimes). In other words, keyphrases written by different humans are lexically similar to the original references, with slightly different phrasings.

\paragraph{Semantic matching is more robust to label variations.} A desired property of keyphrase metrics is outputting consistent scores with similar sets of labels. Using the 200 annotated documents, we compare $F1@M$ and $SemF1$ on the outputs from \textbf{M20} with four sets of labels\footnote{We choose \textbf{M20} as many supervised models' outputs largely overlap with those included in the annotation process.}: the original labels and three sets of newly annotated labels. As shown in Table \ref{tab:metric-stability}, the output of $SemF1$ is more stable using different label sets, indicating a higher robustness and reliability compared to $F1@M$. 

\setlength{\tabcolsep}{3.5pt}
\begin{table}[]
  \centering
  \resizebox{\linewidth}{!}{
  \begin{tabular}{c | c c | c c  }
    \hline
    \multirow{2}{*}{\textbf{Label Sets}}  &   \multicolumn{2}{c|}{\textbf{KP20k}} & \multicolumn{2}{c}{\textbf{KPTimes}} \\
    & $F1@M$ & $SemF1$ & $F1@M$ & $SemF1$ \\
    \hline
    Original & 0.160 & 0.498 & 0.085 & 0.461 \\
    Newly Created 1 & 0.225 & 0.536 & 0.143 & 0.487 \\
    Newly Created 2 & 0.207 & 0.500 & 0.143 & 0.475\\
    Newly Created 3 & 0.254 & 0.556 & 0.128 & 0.455\\
    \hdashline
    \hfil $\sigma$ & 0.039 & \textbf{0.027} & 0.028 & \textbf{0.014} \\
  \hline
  \end{tabular}
  }
  \caption{Evaluating \textbf{M20} with $F1@M$ and $SemF1$ using four label sets. $SemF1$ displays a higher consistency across different label versions, as indicated by a lower standard deviation ($\sigma$).}
  \label{tab:metric-stability}
  
\end{table}

\section{Annotation for Meta-Evaluation}
\label{human-eval-setup}

\paragraph{Annotator Hiring} For all the annotation experiments, we use Amazon Mechanical Turk (MTurk) and designed a qualification task to hire and train annotators. We require the annotators to be located in United States or United Kingdom and have finished more than 1000 HITs with 97\% pass rate. In the qualification task, the annotators are presented with the definition of semantic matching with examples, and then asked to annotate three documents. 46 annotators that have average $\le1.5$ wrong annotations per document are selected. We ensure that the purpose of the main tasks and how we use the annotations are clearly explained in the qualification task to the potential annotators. 

\paragraph{Cost} For reference agreement, a total of 1500 document-level annotations with 13401 phrase-level evaluations are collected from the qualified annotators, costing approximately \$1400. For faithfulness, we collect 6450 phrase-level annotations for KP20k and 6486 annotations for KPTimes, costing approximately \$800. We adjust the unit pay to ensure \$15 hourly pay. 

\begin{figure*}[]
\centering
\fbox{\includegraphics[width=0.9\linewidth]{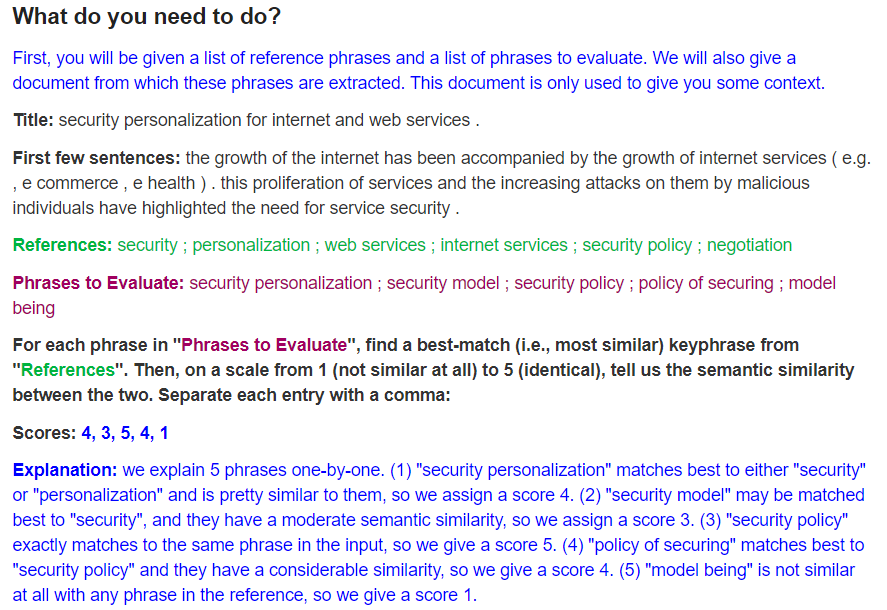}}
\caption{An example of the annotation instructions for the keyphrase evaluation study in \cref{meta-evaluation}.}
\label{scoring-instructions}
\end{figure*}

\begin{figure*}[]
\centering
\fbox{\includegraphics[width=0.9\linewidth]{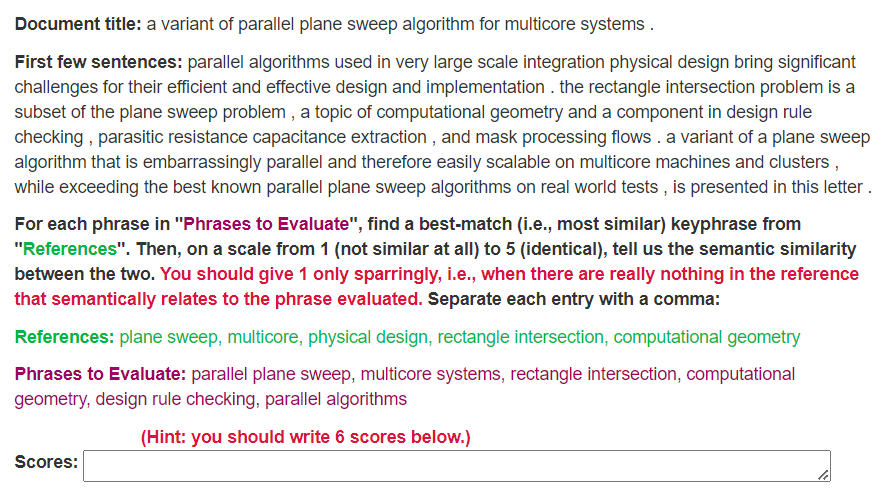}}
\caption{An example of the annotation interface for the keyphrase evaluation study in \cref{meta-evaluation}.}
\label{scoring-interface}
\end{figure*}

\begin{figure*}[]
\centering
\includegraphics[width=0.95\linewidth]{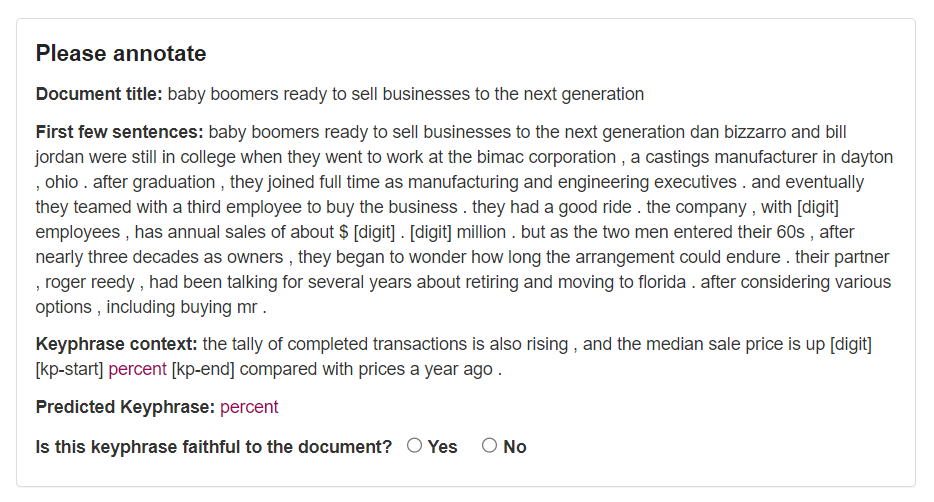}
\caption{An example of the annotation interface for the faithfulness study in \cref{nat-faith-methods}.}
\label{labeling-interface-faithfulness}
\end{figure*}

\paragraph{Interface} The annotation instructions and interface for reference agreement are presented in Figure \ref{scoring-instructions} and Figure \ref{scoring-interface}. The interface for faithfulness is presented in Figure \ref{labeling-interface-faithfulness}.

\section{Qualitative Study}
\label{qualitative-study}

\subsection{$SemF1$ vs. existing metrics}
\label{qual-study-semantic-matching}

In Figure \ref{qualitative-result-matching}, we present a qualitative example where the model (\textbf{M21}) predicts two keyphrases exactly matching to some reference keyphrases and two keyphrases semantically similar to the reference. Correspondingly, human annotators assign partial credits to both of the "near-misses". However, for these phrases, exact matching gives a score 0 and substring matching gives full credit. As the reference and the prediction contain many similar tokens, BertScore is also near 1.0. Semantic matching's scoring is the most similar to humans. 

\subsection{Rethinking model comparisons}
\label{qual-study-model-comparisons}

In Figure \ref{example-zeroshot-outputs-kp20k}, we compare ExHiRD-h and CatSeqTG-2RF1 on two instances from KP20k. When evaluated with exact matching, CatSeqTG-2RF1 is preferred. However, since a lot of correct keyphrases are not recognized by exact matching, the irrelevant concepts and duplications are under-penalized by $F1@M$. By contrast, these issues are identified by the metrics from \textsc{KPEval} dedicated to faithfulness and diversity. With semantic-based evaluation, \textsc{KPEval} suggests that ExHiRD-h outperforms CatSeqTG-2RF1 on these two instances.

\begin{figure}[]
    \centering
    \small
    \resizebox{\linewidth}{!}{
    \fbox{\begin{tabular}{ p{1.1\columnwidth} }
    \textbf{Document Title}: extensional normalisation and type directed partial evaluation for typed lambda calculus with sums \\
    \textbf{Reference Keyphrases}: normalisation, typed lambda calculus, grothendieck logical relations, strong sums \\
    \textbf{Predictions (M21)}: typed lambda calculus, sums, extensional normalisation, grothendieck logical relations \\
    \hdashline
    \textbf{Human}: \\
    \hspace{0.2cm} \textcolor{cyan}{$P=(1.00+0.67+0.60+1.00)/4=$$0.82$} \\
    \hspace{0.2cm} \textcolor{purple}{$R=(0.73+1.00+1.00+0.67)/4=$$0.85$} \\
    \hspace{0.2cm} \textcolor{teal}{$F1=2\times0.85\times0.82/(0.85+0.82)=$$0.83$} \\
    \textbf{Exact Matching}: \textcolor{cyan}{$P=0.50$}, \textcolor{purple}{$R=0.50$}, \textcolor{teal}{$F1=0.50$} \\
    \textbf{Substring Matching}: \textcolor{cyan}{$P=1.00$}, \textcolor{purple}{$R=1.00$}, \textcolor{teal}{$F1=1.00$}\\
    \textbf{BertScore}: \textcolor{cyan}{$P=0.94$}, \textcolor{purple}{$R=0.96$}, \textcolor{teal}{$F1=0.95$}\\
    \textbf{Semantic Matching}: \\
    \hspace{0.2cm} \textcolor{cyan}{$P=(1.00+0.62+0.55+1.00)/4=0.79$} \\
    \hspace{0.2cm} \textcolor{purple}{$R=(0.55+1.00+1.00+0.62)/4=0.79$} \\
    \hspace{0.2cm} \textcolor{teal}{$F1=2\times0.79\times0.79/(0.79+0.79)=0.79$} \\
    \end{tabular}}}
    \caption{An example case from KP20k. Human scores are the average of the scores from three annotators, normalized to a [0,1] range. The small differences in human precision and recall scores are due to annotation noises. Semantic matching's intermediate and final scores are the most similar to human judgments.}
    \label{qualitative-result-matching}
\end{figure}

\begin{figure*}[h!]
\small
\centering
\begin{tabular}{p{0.98\linewidth}}
    \hline
    \textbf{Title:} yet another write optimized dbms layer for flash based solid state storage . \\
    \textbf{Abstract:} flash based solid state storage ( flashsss ) has write oriented problems such as low write throughput , and limited life time . especially , flashssds have a characteristic vulnerable to random writes , due to its control logic utilizing parallelism between the flash memory chips . in this paper , we present a write optimized layer of dbmss to address the write oriented problems of flashsss in on line transaction processing environments . the layer consists of a write optimized buffer , a corresponding log space , and an in memory mapping table , closely associated with a novel logging scheme called incremental logging ( icl ) . the icl scheme enables dbmss to reduce page writes at the least expense of additional page reads , while replacing random writes into sequential writes . through experiments , our approach demonstrated up to an order of magnitude performance enhancement in i o processing time compared to the original dbms , increasing the longevity of flashsss by approximately a factor of two . \\
    \textbf{Reference:} \texttt{icl ; ssd ; incremental logging ; flash memory ; write performance ; database }\\
    \textbf{ExHiRD-h:} \texttt{solid state storage ; flash memory ; write optimized buffer ; incremental logging}  \\
    \textcolor{blue}{\ \ \ \ \ \ \ \ \ -- \textbf{Exact matching:} $F1@M=0.500$ (Present), $F1@M=0$ (Absent) }\\
    \textcolor{purple}{\ \ \ \ \ \ \ \ \ -- \textbf{\textsc{KPEval}:} $SemF1=0.657$, faithfulness$=0.864$, $emb\_sim=0.322$, $RR=1.0$}  \\
    \textbf{CatSeqTG-2RF1:} \texttt{flash based solid state storage ; flash based solid state storage ; incremental logging ; security ; flash memory ; flash memory ; flash memory} \\
    \textcolor{blue}{\ \ \ \ \ \ \ \ \ -- \textbf{Exact matching:}  $F1@M=0.667$ (Present), $F1@M=0$ (Absent) }\\
    \textcolor{purple}{\ \ \ \ \ \ \ \ \ -- \textbf{\textsc{KPEval}:} $SemF1=0.585$, faithfulness$=0.773$, $emb\_sim=0.419$, $RR=0.333$}  \\
    \hline
    \textbf{Title:} bicepstrum based blind identification of the acoustic emission ( ae ) signal in precision turning . \\
    \textbf{Abstract:} it is believed that the acoustic emissions ( ae ) signal contains potentially valuable information for monitoring precision cutting processes , as well as to be employed as a control feedback signal . however , ae stress waves produced in the cutting zone are distorted by the transmission path and the measurement systems . in this article , a bicepstrum based blind system identification technique is proposed as a valid tool for estimating both , transmission path and sensor impulse response . assumptions under which application of bicepstrum is valid are discussed and diamond turning experiments are presented , which demonstrate the feasibility of employing bicepstrum for ae blind identification . \\
    \textbf{Reference:} \texttt{acoustic emissions ; higher order statistics ; blind identification ; precision machining}\\
    \textbf{ExHiRD-h:} \texttt{acoustic emission ; precision turning ; blind system identification ; acoustic emission ; blind source separation}  \\
    \textcolor{blue}{\ \ \ \ \ \ \ \ \ -- \textbf{Exact matching:} $F1@M=0.333$ (Present), $F1@M=0$ (Absent) }\\
    \textcolor{purple}{\ \ \ \ \ \ \ \ \ -- \textbf{\textsc{KPEval}:} $SemF1=0.629$, faithfulness$=0.825$, $emb\_sim=0.122$, $RR=1.0$}  \\
    \textbf{CatSeqTG-2RF1:} \texttt{bicepstrum ; blind identification ; acoustic emission ; precision turning ; diamond turning ; algorithms ; blind source separation ; blind source separation ; blind source separation } \\
    \textcolor{blue}{\ \ \ \ \ \ \ \ \ -- \textbf{Exact matching:}  $F1@M=0.4$ (Present), $F1@M=0$ (Absent) }\\
    \textcolor{purple}{\ \ \ \ \ \ \ \ \ -- \textbf{\textsc{KPEval}:} $SemF1=0.556$, faithfulness$=0.532$, $emb\_sim=0.336$, $RR=0.333$}  \\
    \hline     
\end{tabular}
\caption{A comparison between ExHiRD-h and CatSeqTG-2RF1 on two instances from KP20k. \textsc{KPEval} challenges the results of exact matching with more fine-grained evaluation signals. }
\label{example-zeroshot-outputs-kp20k}
\end{figure*}

\begin{figure*}[ht!]
    \centering
    \small
    \fbox{\begin{tabular}{ p{0.8\linewidth} }
    \textbf{KP20k}: \\
    For each paper, write a short citation text that summarizes some idea reflected in the abstract without copying anything here. Use a fake paper id like [3] or [5] to refer to the paper. Do not present in a summary format. Instead, write as if you are citing the paper in another paper. \\\\

    Title: How Should I Explain? A Comparison of Different Explanation Types for Recommender Systems \\
    Abstract: Recommender systems help users locate possible items of interest more quickly by filtering and ranking them in a personalized way. In particular, we present the results of a user study in which users of a recommender system were provided with different types of explanation. Our study reveals that the content-based tag cloud explanations are particularly helpful to increase the user-perceived level of transparency and to increase user satisfaction even though they demand higher cognitive effort from the user. Based on these insights and observations, we derive a set of possible guidelines for designing or selecting suitable explanations for recommender systems. \\
    Citation: The ability for an artificially intelligent system to explain recommendations has been shown to be an important factor for user acceptance and satisfaction [13].  \\ \\ 
    
    ... two examples omitted ... \\\\ 
    
    Title: [document\_title] \\
    Abstract: [document\_abstract] \\
    Citation:  \\
    \vspace{1pt} \\
    \hdashline
    \vspace{1pt} \\
    \textbf{KPTimes}: \\
    For each piece of news, write several phrases as ad-hoc queries that some people might write if they want to find this article on the Internet. Write your response in 3-5 phrases and separate the phrases with commas. \\
    Title: [document\_title] \\
    Abstract: [document\_abstract] \\
    Citation: \\
    \end{tabular}}
    \caption{Prompts used for instructing GPT-4 to generate the ad-hoc queries for utility evaluation.}
    \label{adhoc-query-prompt}
\end{figure*}

\end{document}